\pdfoutput=1

\documentclass[11pt]{article}

\usepackage[]{ACL2023}

\usepackage{times}
\usepackage{latexsym}
\usepackage{amsmath}
\usepackage{amssymb}
\usepackage{mathtools}
\usepackage{amsthm}
\usepackage{multirow}
\usepackage{blindtext}
\usepackage{tabularray}
\usepackage{longtable}
\usepackage{xltabular}
\usepackage{enumitem}
\usepackage{booktabs}
\usepackage{supertabular}

\usepackage[T1]{fontenc}

\usepackage[utf8]{inputenc}

\usepackage{microtype}

\usepackage{inconsolata}

\title{Towards Uncertainty-Aware Language Agent}

\author{
  Jiuzhou Han$^{{\natural} }$\ \ \ \ \ 
  Wray Buntine$^{{\flat}}$\ \ \ \ \ 
  Ehsan Shareghi$^{{\natural} }$\\
  $^{{\natural} }$~Department of Data Science \& AI, Monash University \\
  $^{{\flat}}$~College of Engineering and Computer Science, VinUniversity\\
  {jiuzhou.han@monash.edu}\ \ \ \ \ \ {wray.b@vinuni.edu.vn} \\{ehsan.shareghi@monash.edu}
}

\begin{document}
\maketitle
\begin{abstract}
While Language Agents have achieved promising success by placing Large Language Models at the core of a more versatile design that dynamically interacts with the external world, the existing approaches neglect the notion of uncertainty during these interactions. We present the Uncertainty-Aware Language Agent (UALA), a framework that orchestrates the interaction between the agent and the external world using uncertainty quantification. Compared with other well-known counterparts like ReAct, our extensive experiments across 3 representative tasks (HotpotQA, StrategyQA, MMLU) and various LLM sizes demonstrate that UALA brings a significant improvement of performance, while having a substantially lower reliance on the external world (i.e., reduced number of tool calls and tokens). Our analyses provide various insights including the great potential of UALA compared with agent fine-tuning, and underscore the unreliability of verbalised confidence of LLMs as a proxy for uncertainty.~\footnote{Our code and data are available at \url{https://uala-agent.github.io}.}
\end{abstract}

\section{Introduction}

Language Agents~\cite{DBLP:journals/corr/abs-2309-02427,DBLP:journals/corr/abs-2309-07870, DBLP:journals/corr/abs-2309-07864} utilise Large Language Models \cite{DBLP:journals/corr/abs-2303-08774,DBLP:journals/corr/abs-2307-09288,DBLP:journals/corr/abs-2305-10403} to interact with the external world (e.g., through tools) and to process collected observations towards solving a task. Having achieved improvements in previously challenging reasoning tasks~\cite{DBLP:conf/iclr/YaoZYDSN023, DBLP:conf/icml/GaoMZ00YCN23, DBLP:journals/corr/abs-2305-17390}, these agents autonomously obtain new knowledge from the world, and leverage memory or self-refinement mechanisms to iteratively improve their reasoning trajectories~\cite{DBLP:journals/corr/abs-2303-11366, DBLP:journals/corr/abs-2305-11738, DBLP:journals/corr/abs-2310-10134}. 

Current language agent designs primarily define the role of an LLM as the planner, moderator, or aggregator of observations and knowledge collected from the external world~(i.e., via tools and APIs), overlooking the potential of LLMs to tackle numerous tasks without depending on any external tool and resource~\cite{DBLP:journals/corr/abs-2303-11366, lu2023chameleon, yin2023lumos, patil2023gorilla}. While exploiting external mechanisms has an advantage in terms of verifiability of outputs (e.g., output extracted from a retrieved Wikipedia article provides a verifiable reference) or improvement on accuracy (e.g., using a Python program to do mathematical calculation instead of relying on the LLM), this design is inherently wasteful in how it utilises the LLM's internal implicit knowledge. In fact, the field of NLP has shown great interest in probing pretrained language models to highlight various types of knowledge they embody directly in their weights (e.g., see \citet{DBLP:conf/emnlp/PetroniRRLBWM19,DBLP:journals/corr/abs-2208-11981,DBLP:conf/emnlp/ShuH0SC23} and references therein). 
 
We postulate that a more effective design for language agents should have a better interplay between the implicit knowledge encoded in LLM's weight and the explicit knowledge of the external world. Drawing upon principles from decision theory, an intelligent agent should possess a reliable mechanism for \emph{measuring uncertainty} of its own proposition or the observations from the environment~\cite{DBLP:journals/ai/BacchusGHK96}. We build on this, and move towards equipping language agents (as a special case of intelligent agents) with the means of measuring uncertainty to efficiently regulate their use of external help (e.g., tools, knowledge bases).

In NLP, in general, there has been a great amount of research on uncertainty estimation for generative tasks such as Machine Translation~\cite{DBLP:conf/icml/OttAGR18}, Summarisation~\cite{DBLP:conf/emnlp/ZablotskaiaPMN023,xu-etal-2020-understanding-neural}, and data-to-text generation~\cite{DBLP:conf/eacl/XiaoW21}. Recently, a dedicated space emerged to study the intersection of LLMs and Uncertainty. This space can be divided into approaches applicable to open-box LLMs which rely on token-level logits~\cite{DBLP:conf/iclr/MalininG21, DBLP:conf/iclr/KuhnGF23}, and those applicable to black-box LLMs leveraging multiple response generation as a proxy for model's confidence~\cite{DBLP:journals/corr/abs-2305-19187}, or relying on LLM's verbal self-awareness of confidence~\cite{DBLP:conf/emnlp/TianMZSRYFM23}. Nonetheless, uncertainty in the context of language agents is an under-explored space, with our work being the first in this direction.

We propose an Uncertainty-Aware Language Agent framework which utilises various existing uncertainty measurement methods as a dynamic switch between LLM's own trajectory or resorting to external resources (i.e., Search Engine, Wikipedia knowledge base) during the course of a reasoning task. We study the properties of this framework on a large space of LLMs  (ChatGPT; \citet{DBLP:journals/corr/abs-2303-08774}, and LLaMA2 70B, 13B, 7B; \citet{DBLP:journals/corr/abs-2307-09288}), and Language Agent frameworks such as ReAct~\cite{DBLP:conf/iclr/YaoZYDSN023}, and CRITIC~\cite{DBLP:journals/corr/abs-2305-11738} on several representative tasks: HotpotQA~\cite{DBLP:conf/emnlp/Yang0ZBCSM18} a free-form QA task; StrategyQA~\cite{DBLP:journals/tacl/GevaKSKRB21} a representative binary QA task; and 57 multiple-choice QA tasks of MMLU~\cite{DBLP:conf/iclr/HendrycksBBZMSS21}. Our key findings and contributions are as follows: 

We propose the first language agent framework which integrates uncertainty in the reasoning trajectories. The proposed design allows for a plug-and-play use of various uncertainty estimation techniques. We highlight across our wide spectrum of experiments that integrating uncertainty not only leads to a significant performance improvement, it also has several practical benefits such as a substantial reduction in external calls and number of output tokens~(i.e., less API charge). Our analysis reveals several additional findings: (1) The gain is significant across different LLM sizes and tasks. (2)  Uncertainties have a degree of calibration because there exists a divergence of uncertainty between correct and incorrect answers. (3) The verbalised confidence of LLMs is an unreliable proxy for uncertainty quantification. (4) Given a limited amount of data, leveraging uncertainty leads to a much higher performance improvement compared to fine-tuning language agents.


\section{Related Work}

\subsection{Language Agents}
The prominent work of ReAct~\cite{DBLP:conf/iclr/YaoZYDSN023} proposes a general language agent framework to combine reasoning and acting (i.e., utilising external tools) with LLMs for solving diverse language reasoning tasks. Reflexion~\cite{DBLP:journals/corr/abs-2303-11366} proposes to use the history of failed trials based on ReAct to do self-reflection, inducing better decision-making in subsequent trials of the language agent. FireAct~\cite{DBLP:journals/corr/abs-2310-05915} fine-tunes language agent with diverse ReAct task-solving trajectories from multiple tasks and prompting methods, and shows having more diverse fine-tuning data can further improve performance of tool-use. CRITIC~\cite{DBLP:journals/corr/abs-2305-11738} allows iterative verification and correction of agent's outputs by interacting with external tools.
\begin{figure*}[t]
    \centering
    \includegraphics[scale=0.95]{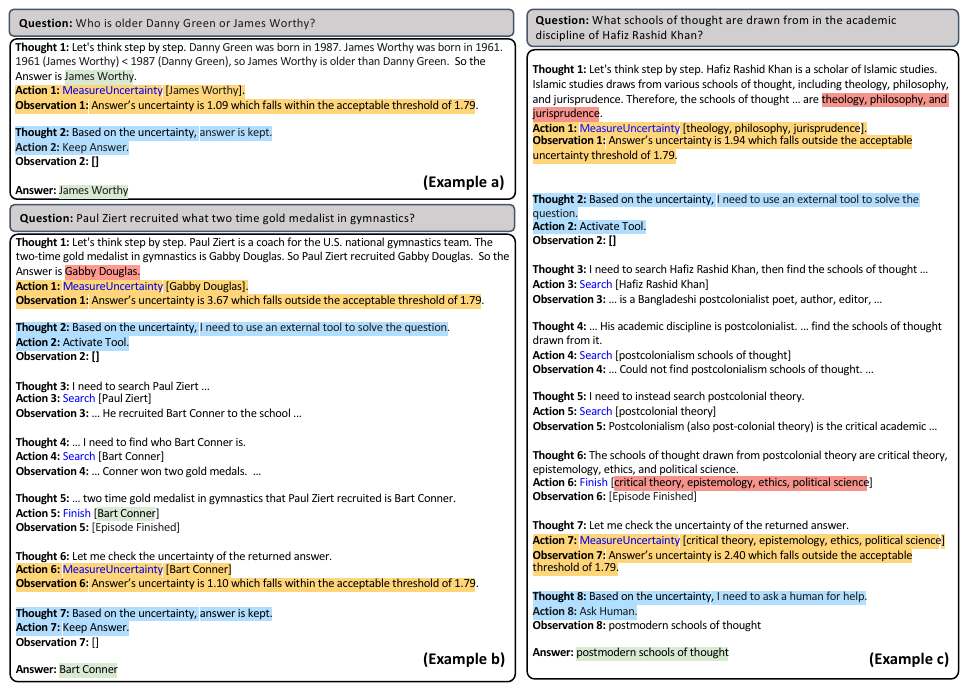}
    \caption{Examples of single-inference UALA trajectories: (a) illustrates the trajectory where CoT answer falls inside the certainty region. (b) is the trajectory where CoT is too uncertain and tool is activated, arriving at a final response which falls in the acceptable certainty region (denoted by UALA-S).  (c) is the trajectory where both CoT and tool-generated responses are  uncertain, and the agent asks for help from human (denoted by {UALA-S+Oracle}).}
    \label{fig:example-singleinference}
\end{figure*}

 As an earlier work in this direction, Toolformer~\cite{DBLP:journals/corr/abs-2302-04761} attempts to use different APIs (e.g., Wikipedia, QA, Calculator) in a self-supervision manner and autonomously choose proper APIs to use when solving different tasks. \citet{DBLP:journals/corr/abs-2305-13068} propose a two-stage framework which allows the model to learn through feedback derived from tool execution. Gorilla~\cite{patil2023gorilla} is a retrieval-aware system fine-tuned on open-source LLMs for enabling LLMs to interact with tools, which enhances the LLM’s ability to accurately identify the appropriate API for a specific task. Lumos~\cite{yin2023lumos} is a language agent fine-tuned on open-source LLMs which contains three modules~(Planning Module, Grounding Module and Execution Module) for solving complex interactive tasks. Chameleon~\cite{lu2023chameleon} is a plug-and-play compositional reasoning framework, which infers the appropriate sequence of tools to compose and execute in order to solve the problems. PoT~\cite{DBLP:journals/corr/abs-2211-12588} and PAL~\cite{DBLP:conf/icml/GaoMZ00YCN23} utilise LLMs to generate programs and leverage external program interpreter to execute for arithmetic and symbolic reasoning tasks.

None of these frameworks consider any role for uncertainty in the agent's interactions with the external world. Next, we present a brief overview of the intersection of uncertainty and LLMs.

\subsection{Uncertainty in Generation with LLMs}

A common practice in leveraging uncertainty during language generation is via sampling or decoding approaches which do not measure the uncertainty directly, and instead rely on the stochasticity over the prediction space along with a form of  aggregation approach such as majority voting. Self-Consistency~\cite{DBLP:conf/iclr/0002WSLCNCZ23} is a widely known sampling method, which takes majority voting over multiple sampling outputs. Since it is based on exact matching, it is most useful in tasks for mathematical reasoning or fact-based question answering tasks. Minimum Bayes-Risk Decoding~\cite{kumar2004minimum} is also based on multiple candidate outputs sampled from a model, and it selects the candidate with lowest expected risk.  \citet{suzgun-etal-2023-follow} highlight the connection between Self-Consistency and MBR and showcase the benefit of MBR using a soft alignment function (i.e., semantic similarity) in tasks involving open-ended, longer free-form text generation (e.g., summarisation, translation). For a  comprehensive review of sampling and decoding methods in NLG, we refer the readers to \citet{wiher-etal-2022-decoding}.

We focus on leveraging uncertainty estimation in free-form QA with short answers. Uncertainty estimation in free-form NLG remains a challenge for LLMs due to the diversity of the outputs. The recent research on this area can be classified as:

\noindent\textbf{Logits- or Entropy-based Methods.}
\citet{DBLP:conf/iclr/MalininG21} calculate the accumulative predictive entropies over multiple generations to estimates the uncertainty. But this method ignores the semantic equivalence phenomenon in generated texts. \citet{DBLP:conf/iclr/KuhnGF23} propose semantic entropy to solve the difficulty of semantic equivalence in uncertainty estimation. 
FLARE~\cite{DBLP:conf/emnlp/JiangXGSLDYCN23} actively decides when and what to retrieve from external knowledge resources based on the token probabilities across the generation. \citet{DBLP:journals/corr/abs-2310-04782} implement multiple uncertainty calculation methods based on the token-level logit output. \citet{DBLP:journals/corr/abs-2307-01928} utilise uncertainty alignment of LLM-based planners in solving complex step-by-step planing problems for robots via conformal prediction. 

\noindent\textbf{Prompt-based Methods.}
\citet{DBLP:conf/emnlp/TianMZSRYFM23, DBLP:journals/corr/abs-2207-05221, DBLP:journals/tmlr/LinHE22} prompt LLMs to directly express the confidence or uncertainty along with their generations. \citet{DBLP:journals/corr/abs-2305-19187} propose to estimate the uncertainty based on the similarities of multiple generations. We refer the readers for a more comprehensive overview to read ~\citet{fadeeva-etal-2023-lm} and references therein.

We compare both types of methods.\footnote{Since our focus is not on improving the performance of the uncertainty estimation methods, we didn't aim for solving the potential issues (i.e., semantic equivalence, inequality biases) when using logits-based methods.}

\section{Uncertainty-Aware Language Agent}
\label{sec:unc}
We present Uncertainty-Aware Language Agent, \textsc{Uala}, a framework that integrates uncertainty in language agent's cycle of Thought, Action, and Observation. The uncertainty moderates the interaction between the LLM and the external world, facilitating a more effective and efficient dynamic. Given an input question, the agent's potentially goes through 3 modes of attempt, each involving a thought, an action, and an observation (similar to ReAct). Figure~\ref{fig:example-singleinference} presents real trajectories for each mode of attempt. In our approach, as illustrated in Example (a), we initially generate a response to a query using standard or Chain-of-Thought (CoT;~\citet{DBLP:conf/nips/Wei0SBIXCLZ22}) prompting. This response is then assessed by an uncertainty estimation module, which determines the subsequent action between two courses: either accepting the response as-is, or employing external tools for a more certain solution, as shown in Example (b). This process is similarly applied when using tool-activated responses. If the uncertainty remains high post-tool engagement, the agent seek assistance from an external source such as a human, as depicted in Example (c). This protocol ensures a more robust and reliable response generation by actively addressing uncertainty at each step. Next, we describe our uncertainty estimation mechanisms.
\subsection{Uncertainty Estimation}
Uncertainty estimation methods are broadly categorised into two types: single-inference based and multi-inference based. Single-inference uncertainty estimation calculates the uncertainty based on one output, necessitating access to the token log-probabilities within that output. Multi-inference uncertainty estimation computes the uncertainty based on a set of outputs, eliminating the need for individual token log-probabilities.
\subsubsection{Single-inference Uncertainty Estimation}\label{sec:free-form}
The methods employed vary based on the answer being a single-token (e.g., yes or no) or free-form (multi-token) format. 

\noindent\textbf{Free-form Answer.}
Given an output (i.e., Thought, Observation) containing an answer, the answer $Y$ comprising $n$ tokens $[y_1, y_2, \ldots, y_n]$ along with their corresponding token log-probabilities $[p_1, p_2, \ldots, p_n]$ is extracted.Following the previous work~\cite{DBLP:journals/corr/abs-2310-04782}, we first apply a softmax function to the token log-probabilities to obtain $[z_1, z_2, \ldots, z_n]$. Subsequently,  we explore five methods of calculating the uncertainty $u$ based on the probabilities (see Table~\ref{tab:free-form} for a summary).

\noindent\textbf{Single-token Answer.}
For single-token answers, applying a softmax function to their log-probability is not meaningful due to the presence of only one token. Instead, the uncertainty is determined directly by taking the absolute value of the token's log-probability. In this scenario, a higher probability of the token (approaching 1) results in its log-probability nearing 0, indicating low uncertainty. Conversely, a lower token probability signifies higher uncertainty. Therefore, for a given answer token 
$y$ with log-probability $p$, the uncertainty is defined as $u=|p|$.

\begin{table}[t]\footnotesize
\centering
\begin{tabular}{ll}
\toprule
\textbf{Method}              & \textbf{Formula}                                             \\ \hline
Minimum                    & \( u=-\log(\min(z_1, z_2, \ldots, z_n)) \)                  \\
Average                    & \( u=-\log(\text{Avg}(z_1, z_2, \ldots, z_n)) \)            \\
Normalised Product         & \( u=-\log(z_1 \times z_2 \times \ldots \times z_n)^{\frac{1}{n}} \) \\
Log-sum                    & \( u=-\sum_{i=1}^n \log(z_i) \)                             \\
Entropy                    & \( u=-\sum_{i=1}^n z_i \cdot \log(z_i) \)                   \\\bottomrule
\end{tabular}
\caption{Five methods of calculating the uncertainty $u$ of a free form output of length $n$.}
\vspace{-2mm}
\label{tab:free-form}
\end{table}
\subsubsection{Multi-inference Uncertainty Estimation}
Multi-inference uncertainty estimation calculates the uncertainty of an answer based on multiple outputs from an LLM. This process begins with obtaining a primary answer, $a_*$, using the LLM's greedy decoding method. Subsequently, the LLM is prompted to sample $n=9$ additional answers with temperature $0.7$, forming answers set $A=\{a_1, a_2, \ldots, a_n\}$. The uncertainty of the primary answer is then calculated based on how frequently it appears within set $A$. Essentially, the less frequently $a_*$ occurs in $A$, the higher the uncertainty attributed to it, and vice versa. Specifically,
\begin{equation}
\setlength\abovedisplayskip{1pt}%
    \setlength\belowdisplayskip{2pt}%
    \setlength\abovedisplayshortskip{1pt}%
    \setlength\belowdisplayshortskip{2pt}%
    u=\frac{1}{n}\sum_{i=1}^n \delta \left(a_i, a_*\right),
\end{equation}

where $\delta \left(a_i, a_*\right)$ is an indicator function that equals to 1 if $a_i\neq a_*$, and 0 otherwise.

\subsubsection{Uncertainty Threshold}\label{sec:sin-inf-thre}
The decision to accept an answer or resort to alternative mechanisms hinges on the level of uncertainty associated with that answer. To this end, we propose different ways of setting the uncertainty threshold for single-inference and multi-inference uncertainty estimation.

    For \textbf{Single-inference} setting, we adopt a subset of the training data to create a calibration set. We first use standard prompting or CoT prompting to query an LLM to get the answers and collect the correctly answered questions as the calibration set. {We show this subset could be as small as 200 examples (see Appendix~\ref{subsec:calib-size}).} We use a single-inference uncertainty estimation method to calculate the uncertainty of each answer in the calibration set. The same single-inference uncertainty estimation method is used for the calibration set and test set. Based on the answer uncertainty in the calibration set, we apply three methods to estimate the threshold.\footnote{For cost reason, we apply the same estimation when deciding to accept the LLM (as in Example a) or Tool-activated responses (as in Example b and c of Figure~\ref{fig:example-singleinference}).} Specifically, the Max method utilises the maximum uncertainty in the calibration set as the threshold, while the Mean uses the average uncertainty, and the last approach uses a Quantile~\cite{DBLP:journals/corr/abs-2307-01928} value of uncertainty in calibration set as the threshold. For the estimation of uncertainty threshold in \textbf{Multi-inference} setting, we adopt the same subset of the training data as the calibration set. We use the average uncertainty of the answers in the calibration set as the threshold.

Figure~\ref{fig:example-singleinference} provides examples of the UALA trajectories. After obtaining the uncertainty threshold, for an answer where the uncertainty is above the threshold, alternative modes are activated (i.e., Tool Activation as in Example b, or asking from human if tool-use result is still uncertain as in Example c). 

\section{Experiments}
We explore three types of QA tasks
and utilise Wikipedia and Google Search and external tools:

\noindent\textbf{HotpotQA}~\cite{DBLP:conf/emnlp/Yang0ZBCSM18} is a challenging QA dataset which requires multi-hop reasoning over some Wikipedia passages. The answer of the questions in HotpotQA is free-form with an arbitrary length. We randomly select 500 questions from the training set for the creation of calibration set, and 500 questions from the dev set for evaluation using the same random seed as ReAct~\cite{DBLP:conf/iclr/YaoZYDSN023}. \textbf{StrategyQA}~\cite{DBLP:journals/tacl/GevaKSKRB21} is an open-domain QA dataset which requires implicit reasoning steps. It only contains binary answers (yes or no). We randomly select 500 questions from the training set to create the calibration set. For the evaluation, we use the same dev set used in FireAct~\cite{DBLP:journals/corr/abs-2310-05915} containing 229 questions. \textbf{MMLU}~\cite{DBLP:conf/iclr/HendrycksBBZMSS21} is a multi-choice QA dataset in various domains which covers 57 tasks including elementary mathematics, US history, computer science, law, and more. This dataset is to evaluate model’s academic and professional understanding. Each question contains four options and only one option is correct. To cover all different topics, we randomly sample 10 questions for each task from dev set to create the calibration set, and similarly sampled 10 question for each task (total of 570 questions) from test set for evaluation.

\noindent\textbf{Wikipedia.}
Following the Wikipedia web API setting used in ReAct~\cite{DBLP:conf/iclr/YaoZYDSN023}, the tool prompt contains three actions. (1) \textbf{search[entity],} which returns the initial five sentences from the entity wiki page if it exists. Otherwise, it suggests the top five similar entities from the Wikipedia search engine. (2) \textbf{lookup[string]}, which functions akin to the Ctrl+F in a browser, returning the subsequent sentence on the page containing the specified string. (3) \textbf{finish[answer]}, which contains an answer and finishes the current task. This tool is used to retrieve some relevant knowledge of the entities in the query, which helps the agent to do the reasoning based on the external knowledge and its internal knowledge. The Wikipedia tool is used for HotpotQA and StrategyQA datasets. The universality of \textbf{Google Search.} is better than Wikipedia. It can search any query which is suitable for more complex reasoning tasks. Therefore, we adopt Google Search tool for MMLU task. Following FireAct~\cite{DBLP:journals/corr/abs-2310-05915}, the Google search tool we use is based on SerpAPI\footnote{\url{https://serpapi.com}}. The tool prompt contains two actions: (1) \textbf{search[query]}, which searches a query on Google and returns a short snippet containing the answer of the query. The returned snippet is the first existent item from “answer box”, “answer snippet”, “highlight words”, or “first result snippet”, which ensures the response is short and relevant. (2) \textbf{finish[answer]}, which returns the answer and finishes the episode. 

\subsection{Baselines and Experimental Setup}\label{sec:baselines}
\textbf{Baselines.} We conduct experiments on ChatGPT\footnote{\url{https://openai.com/blog/gpt-3-5-turbo-fine-tuning-and-api-updates}} and LLaMA2~\cite{DBLP:journals/corr/abs-2307-09288}. For ChatGPT, since we need access to the token log-probability for uncertainty estimation, we use \texttt{gpt-3.5-turbo-instruct} which is a Completion API. For LLaMA2, we adopt LLaMA2-70B as our main backbone, and also compare the performance of different sizes of LLaMA2 (7B and 13B) in Section~\ref{sec:LLaMA2-dif-size}. We compare UALA with \textbf{Standard}~\cite{DBLP:conf/nips/BrownMRSKDNSSAA20}, \textbf{Chain-of-Thought} (CoT; \citet{DBLP:conf/nips/Wei0SBIXCLZ22})\footnote{We use ``Let's think step by step.'' \cite{kojima2022large}.}, \textbf{Self-Consistency} (SC; \citet{DBLP:conf/iclr/0002WSLCNCZ23})\footnote{Generates multiple outputs using a sampling method and takes the majority result as the final answer. For the sampling method, we use temperature $0.7$ to generate 9 samples.}, \textbf{ReAct}\cite{DBLP:conf/iclr/YaoZYDSN023}\footnote{Combines the CoT with actions to retrieve relevant knowledge from external tools, synergizing reasoning and acting.}, and \textbf{Backoff} which backs off to response by Standard or CoT when the Tool Activation trajectories do not arrive at an answer within a given number of steps. For UALA, when the result includes human oracle help, it is marked by \textbf{+Oracle}. We also compare with the \textbf{CRITIC}~\cite{DBLP:journals/corr/abs-2305-11738} in Section~\ref{sec:critic}. Prompts and few-shot examples are in Appendix~\ref{appd:prompt_template}.

\begin{table*}[t]\footnotesize
\centering
    \resizebox{\textwidth}{!}{%
\begin{tabular}{clcccccc}
\toprule
\multicolumn{2}{c}{\multirow{2}{*}{\textbf{Methods}}}    & \multicolumn{3}{c}{\textbf{ChatGPT}}        & \multicolumn{3}{c}{\textbf{LLaMA2-70B}}       \\ \cmidrule(lr){3-5}\cmidrule(lr){6-8}
\multicolumn{2}{c}{}                            & \textbf{HotpotQA}  & \textbf{StrategyQA} & \textbf{MMLU}      & \textbf{HotpotQA}  & \textbf{StrategyQA} & \textbf{MMLU}       \\ \hline
\multirow{5}{*}{Baselines} & Standard           & 29.8(0)   & 57.6(0)    & 69.0(0)   & 30.0(0)   & 65.9(0)    & 64.7(0)   \\
                           & CoT                & 34.8(0)   & 55.9(0)    & 49.1(0)   & 35.6(0)   & 63.8(0)    & 39.3(0)   \\
                           & Self-Consistency   & 39.4(0)   & 58.5(0)    & 70.0(0)   & 37.4(0)   & 67.7(0)    & 67.2(0)   \\
                           & ReAct              & 32.0(2,114) & 55.5(709)  & 55.8(1,824) & 32.4(2,094) & 58.1(890)  & 30.7(2,808) \\
                           & ReAct+Backoff      & 35.4(2,114) & 61.6(709)  & 59.8(1,824) & 37.0(2,094) & 66.8(890)  & 62.8(2,808) \\ \hline
\multirow{6}{*}{Ours}      & UALA-S         & 38.2(403)  & 65.5(134)   & 69.8(662) & 36.4(350)  & 69.0(298)   & 56.7(1354) \\
                           & UALA-S+Backoff & 39.2(403)  & 66.4(134)   & 71.4(662) & 37.3(350)  & 71.6(298)   & 69.8(1354) \\
                           & UALA-S+Oracle & 41.4(403)  & 67.5(134)   & 75.8(662) & 42.2(350)  & 80.1(298)   & 70.5(1354) \\
                           & UALA-M         & 39.8(1,183) & 63.8(234)   & 67.9(641) & 38.4(925) & 66.1(572)  & 56.7(1,196) \\
                           & UALA-M+Backoff & \textbf{41.3(1,183)} & \textbf{66.9(234)}   & \textbf{72.2(641)} & \textbf{40.2(925)} & \textbf{71.8(572)}  & \textbf{70.3(1,196)} \\
                           \bottomrule
\end{tabular}
}
\caption{Results of three question answering datasets on ChatGPT and LLaMA2-70B. The metric is exact match (EM) and the number in the bracket represents the number of tool calls. UALA-S denotes using single-inference uncertainty estimation method and UALA-M denotes using multi-inference uncertainty estimation method.}
\label{tab_main_results}
\end{table*}
\textbf{Experimental Setup.} In all our experiments, unless stated otherwise, we use Entropy for free-form single-inference uncertainty estimation for HotpotQA and 0.9 Quantile for single-inference uncertainty threshold estimation, which perform best in the training set (we demonstrate the effect of various uncertainty estimation and threshold methods in Appendix~\ref{sec:sin-es-me} and Quantile value in Section~\ref{sec:quantile_vale}). We compare different free-form single-inference uncertainty estimation methods in Section~\ref{sec:sin-es-me}. The multi-inference uncertainty estimation is based on 9 samples generated using temperature $0.7$. For HotpotQA, we use 6-shot learning and for StrategyQA and MMLU we use 4-shot learning. For HotpotQA, since CoT performs better than Standard, we use CoT as the base prompt method. For StrategyQA and MMLU, we use Standard as the base prompt method. Following the previous work~\cite{DBLP:journals/corr/abs-2310-05915, DBLP:conf/iclr/YaoZYDSN023}, we evaluate using the exact match (EM). In addition, we also report the total number of tool calls in the bracket following the EM. 


\subsection{Main Results}
 Our main results are demonstrated in Table~\ref{tab_main_results} (for relative improvement details, refer to Table~\ref{tab:result_statistic} of Appendix).  
Some key observations:

(1) CoT outperforms Standard on HotpotQA, while Standard excels on StrategyQA and MMLU. Self-Consistency consistently enhances results across three datasets and two LLMs. ReAct, when used for every instance, underperforms Standard/CoT/Self-Consistency. With the integration of backoff, ReAct+Backoff shows improvement but is generally  behind Self-Consistency, highlighting the benefit of SC's sampling and majority voting as a proxy for capturing uncertainty.

(2) UALA-S significantly betters ReAct's performance, cutting tool use by over half, and surpasses Standard/CoT across all datasets. UALA-M achieves similar performance to UALA-S but with increased tool use. UALA-S+Backoff outperforms ReAct+Backoff and often exceeds Self-Consistency. UALA-M+Backoff delivers the best results in all settings on three datasets.

(3) The largest gain in improvement by UALA is observed for HotpotQA (free-form), followed by StrategyQA (binary), and MMLU (multiple choice). This is expected as the free-form response space is much larger and diverse, compared with MCQ type of questions. In Section~\ref{sec:corr_vs_incorr_uncertainty}, we highlight the difference in gain could be explained in terms of the amount of uncertainty divergence between correct and incorrect answers in each task.

(4) The average (single-inference and multi-inference) EM improvement for ChatGPT with LLaMA2-70B compared to Standard/CoT results: ChatGPT gains 11.7\% and LLaMA2-70B gains  8.9\%. This could be an indication that ChatGPT is likely to produce better-calibrated probability estimates, leading to a more reliable uncertainty estimation on training set that generalises to test set. This could be an artefact of the two models' difference in size and training protocol. For further analysis of the inference cost see Section~\ref{sec:inf-cost}. 

(5) The results from UALA-S+Oracle underscore an additional aspect of the value of uncertainty. This feature is particularly crucial in sensitive domains, as it can deter the agent from generating incorrect responses. Instead of risking an erroneous answer, the agent defers to human (we simulate this by using gold answer) when the response uncertainty is still high after tool activation.


\subsection{Correct vs. Incorrect Answer Uncertainty}
\label{sec:corr_vs_incorr_uncertainty}
\begin{figure}[t]
    \centering
    \includegraphics[trim={0.1cm 0.1cm 0.1cm 0.1cm},clip,scale=0.35]{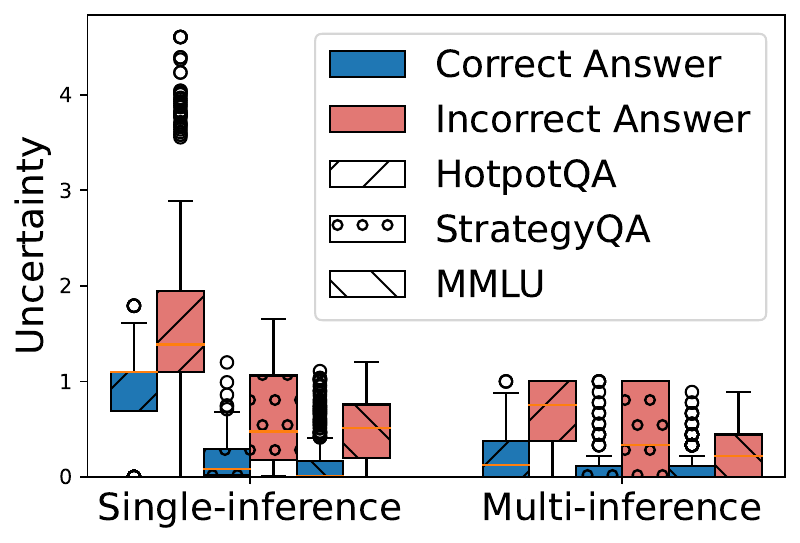}
     \caption{The boxplots of uncertainty range for correct and incorrect answers of three datasets on ChatGPT.}
     \vspace{-2.5mm}
     \label{tab:answer_uncertainty}
\end{figure}

Figure~\ref{tab:answer_uncertainty} illustrates the uncertainty visualisation or single-inference and multi-inference answers on HotpotQA (500, dev set), StrategyQA (229, dev set), and MMLU (570, test set) with ChatGPT. In both single-inference and multi-inference settings, correct answers consistently exhibit lower uncertainty compared to incorrect ones. This difference is statistically significant (see Table~\ref{table:stat_comparison} of Appendix). When calculating the difference between the average uncertainty of correct and incorrect answers we observe the largest difference to belong to HotpotQA, followed by StrategyQA, and MMLU. This explains why the gain from UALA follows the same pattern in the main results.

\subsection{Effect of LLM Size}
\label{sec:LLaMA2-dif-size}
We experimented with HotpotQA using LLaMA2-7B and LLaMA2-13B, employing the same settings followed for the LLaMA2-70B experiments. As reported in Table~\ref{tab:diff-size-LLaMA2}, we observe a similar trend to our main results. It is noteworthy that the gain is sensitive to the backbone LLM size. For example, comparing ReAct and UALA-S, the gain is 35\% on 7B, shrinks to 29\% for 13B, and then to 19\% for the 70B. This is anticipated as larger model capacity could allow for a more confident learning of the data distribution during training.

\subsection{Inference Cost}\label{sec:inf-cost}
\begin{figure}[t]
    \centering
    \includegraphics[trim={0.2cm 0.3cm 0cm 0cm},clip,scale=0.25]{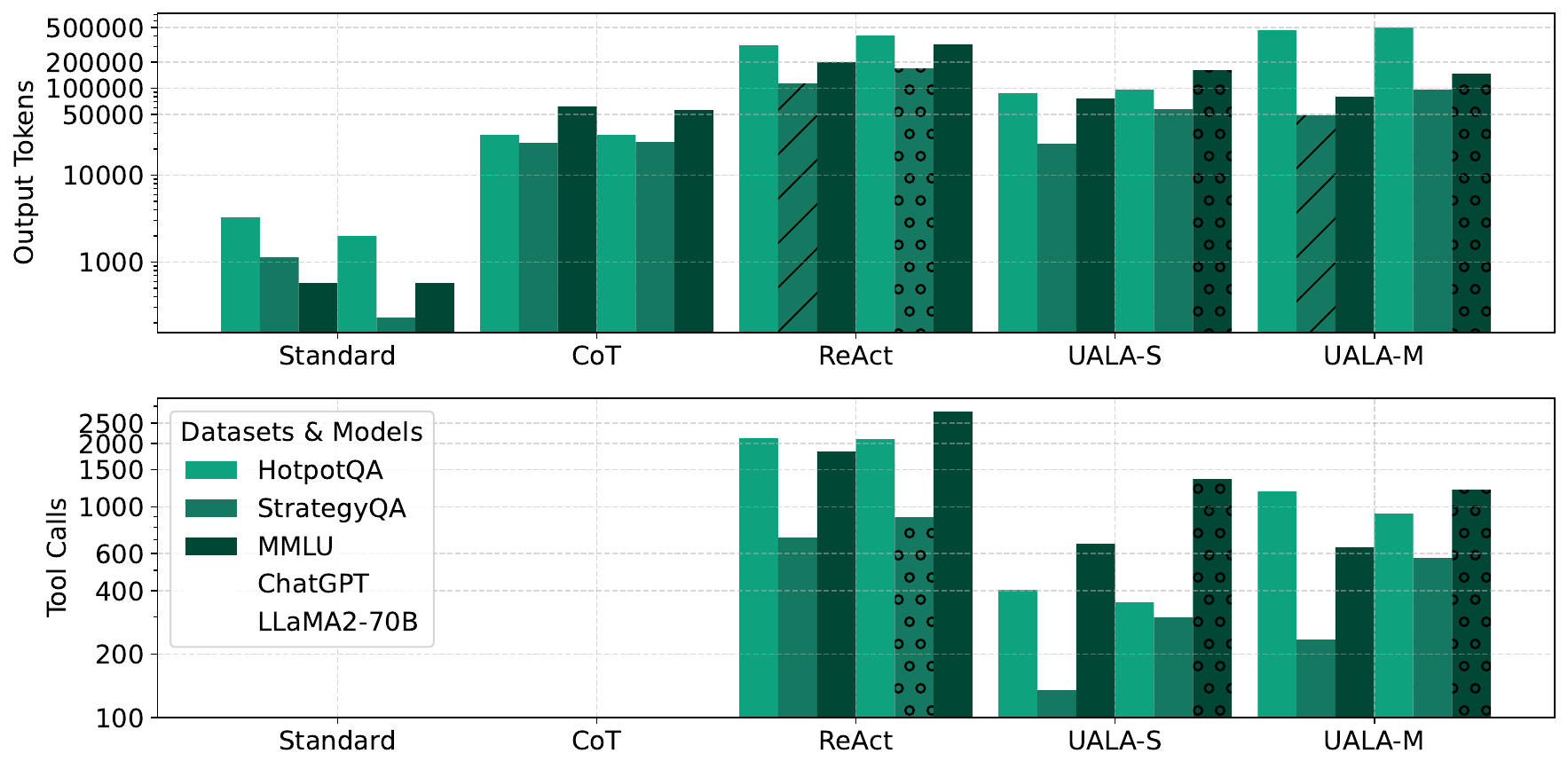}
    \vspace{-2mm}
    \caption{The number of output tokens and tool calls.}
    \label{fig:cost_token_tool}
\end{figure}

Figure~\ref{fig:cost_token_tool} reports the number of output tokens and tool calls per method. ReAct consumes substantially more ($5\times$) output tokens than CoT. Compared with ReAct, UALA-S reduces the number of output tokens by more than 65\%. UALA-M consumes more output tokens as it relies on multiple inference. Both UALA methods can substantially reduce tool calls more than 50\% compared with ReAct. Table~\ref{tab:inf_cost} demonstrates the average inference time per instance for different methods and LLMs on HotpotQA. Standard and CoT do not involve an external tool call, hence faster inference time. The UALA-S, given its selective tool call, has a much lower inference time compared with ReAct. These figures highlight a practical benefit of using uncertainty to reduce the number of token usage and tool calls, along with a significant gain in performance.

\begin{table}[t]
\centering
\resizebox{1\columnwidth}{!}{%
\begin{tabular}{lccccccccc}
\toprule
& \rotatebox{90}{Standard} & \rotatebox{90}{CoT} & \rotatebox{90}{SC} & \rotatebox{90}{ReAct} & \rotatebox{90}{+Backoff} & \rotatebox{90}{UALA-S} & \rotatebox{90}{+Backoff} & \rotatebox{90}{UALA-M} & \rotatebox{90}{+Backoff} \\
\midrule
LLaMA2-7B & 21.2 & 21.8& 22.0 & 16.8 & 23.8 & 22.8 & 24.2 & 22.6  & \textbf{24.6} \\
& (0) & (0) & (0) & (2,324) & (2,324) & (479) & (479) & (1,386) & \textbf{(1,386)} \\ \hline
LLaMA2-13B & 23.4 & 24.8  & 26.4 & 19.8& 25.0 & 25.6  & 27.3 & 26.5 & \textbf{28.4} \\
& (0) & (0) & (0) & (2,371) & (2,371) & (539) & (539) &  (1,525) & \textbf{(1,525)} \\
\bottomrule
\end{tabular}}
\caption{HotpotQA Results on LLaMA2-7B and 13B.}
\vspace{-2mm}
\label{tab:diff-size-LLaMA2}
\end{table}

\subsection{Verbalised Uncertainty Self-Awareness}
A recent work~\cite{DBLP:conf/emnlp/TianMZSRYFM23} reports that RLHF-LMs~(e.g., ChatGPT) are able to directly verbalise well-calibrated confidence score along with the predicted answer by prompting. This uncertainty estimation method can also be used in UALA. To verify this idea, we use their prompt templates modified for our task (See Appendix~\ref{sec:verbalised_template}) to generate the answer with a corresponding confidence score ranging from 0 to 1. When the confidence of an answer is low (i.e., uncertainty is high), we follow the similar protocol to our approach. For the confidence threshold, we demonstrate the results of using threshold value from 0.1 to 0.9. The result of HotpotQA on ChatGPT is shown in Table~\ref{fig:confi_plot}. When using confidence threshold value below 0.8, both of these two methods achieve better results with a small number of tool-use than baselines, but the improvement still falls behind our uncertainty-estimation methods (in Table~\ref{tab_main_results}). We observe that when the model verbalises confidence, there is a large number of answers with confidence above 0.9, leading to degradation of performance and increase of tool-use with 0.9 threshold. This observation underscores that despite the accessibility of expressed confidence of LLMs, in practice it remains as an unreliable indicator of answer uncertainty.

\begin{table}[t]\footnotesize
\centering
\resizebox{0.95\columnwidth}{!}{%
\begin{tabular}{lcccc}
\toprule
\multicolumn{1}{c}{}                                 & \textbf{Standard} & \textbf{CoT} & \textbf{ReAct} & \multicolumn{1}{l}{\textbf{UALA-S}} \\ \hline 
ChatGPT                                            & 0.5s/it           & 1s/it        & 12s/it         & 3s/it                                 \\
LLaMA2-70B                                           & 50s/it            & 50s/it       & 180s/it        & 70s/it                                \\
LLaMA2-13B                                           & 25s/it            & 25s/it       & 120s/it        & 45s/it                                \\
LLaMA2-7B                                            & 20s/it            & 20s/it       & 100s/it        & 35s/it                               \\ \bottomrule
\end{tabular}%
}
\caption{The average inference time per instance (seconds/iteration) of different methods for HotpotQA. The inference of LLaMA2 is done on a single A40 GPU.}
\label{tab:inf_cost}
\vspace{-3.5mm}
\end{table}
\begin{table}[t]
    \centering
    \resizebox{\columnwidth}{!}{%
    \begin{tabular}{lccccc}
    \toprule
    Confidence    &0.1& 0.3&0.5&0.7&0.9\\\hline
  Verbal&37.2 (29)&37.4 (69)&37.2 (87)&37.2 (97)&34.4 (1,109)\\
+Backoff&37.2 (29)&37.5 (69)&37.3 (87)&37.3 (97)&36.8 (1,109)\\
\bottomrule 
    \end{tabular}}
    \caption{The results of HotpotQA on ChatGPT using LLM's verbalised confidence \cite{DBLP:conf/emnlp/TianMZSRYFM23} instead of uncertainty estimation.} 
    \label{fig:confi_plot}
\end{table}
\begin{table}[t]\footnotesize
\centering
\resizebox{\columnwidth}{!}{%
\begin{tabular}{llccc}
\toprule
\textbf{Tasks}&\textbf{Methods}                    &\textbf{Training Size}& \textbf{ChatGPT} & \textbf{LLaMA2-70B} \\ \hline
\multirow{4}{*}{HotpotQA}&FireAct &162 &        27.8     & 27.8                \\
&FireAct &512 &     33.8       & 30.0              \\
&ReAct& No fine-tuning&32.0             & 32.4                \\
&UALA-S & No fine-tuning& 38.2             & 36.4                \\\hline
\multirow{4}{*}{StrategyQA}&FireAct & 283 &      60.7    &     63.8            \\
&FireAct & 567 &      64.9      &      64.6         \\
&ReAct& No fine-tuning&   55.5          &           58.1      \\
&UALA-S & No fine-tuning&   65.6           &   69.0              \\
\bottomrule
\end{tabular}}
\caption{Results of FireAct vs. UALA-S on HotpotQA and StrategyQA. The ReAct and UALA-S results are based on 6-shot and the off-the-shelf LLM backbones.}
\label{tab:finetuned_reslts}
\vspace{-2mm}
\end{table}
\subsection{Language Agent Fine-tuning vs. UALA}
We demonstrate the comparison between UALA-S and fine-tuning language agents following the FireAct setting \cite{DBLP:journals/corr/abs-2310-05915}. For ChatGPT, we use the official GPT-3.5-Turbo fine-tuning API; for LLaMA2-70B, we use LoRA~\cite{hu2022lora}. 
To have a side-by-side comparison, we use the same 500 training samples used for the calibration set, to construct the fine-tuning data. Mimicking the FireAct setting, we ran the 500 examples using ReAct with ChatGPT, and collected the successful trajectories as the training data for FireAct. This amounted to 162 training examples for HotpotQA and 283 for StrategyQA. In addition, to match the amount of training data as FireAct setting, we also ran an additional 1000 examples to increase the amount of successful training trajectories to 512 for HotpotQA and 567 for StrategyQA. The results of supervised fine-tuning of Language Agents instances are presented in Table~\ref{tab:finetuned_reslts}.  

Interestingly, on HotpotQA using 162 training examples, FireAct under-performs the few-shot (6-shots) ReAct agent, while it outperforms ReAct on StrategyQA using 283 training examples. Increasing the amount of training data to 500+ leads to improvement on both LLMs with fine-tuned ChatGPT-based agents outperforming the ReAct counterpart on both datasets. UALA-S achieves the best result without any fine-tuning and using only the 500 samples for creating the calibration set. This capitalises an obvious empirical advantage for utilising uncertainty instead of fine-tuning in the presence of small amount of data. Fine-tuning still has its own merit in the presence of large training data. As highlighted by~\citet{DBLP:journals/corr/abs-2310-05915}, smaller open-source LMs could potentially catch up with stronger LMs on a particular task given enough fine-tuning data.

\section{Conclusion}
We present the Uncertainty-Aware Language Agent (UALA), which improves a language agent's efficiency in interaction with the external world to answer various QA problems. We first measure the uncertainty of an agent on a small set of examples and establish an acceptable uncertainty threshold. Then the uncertainty estimation function is integrated into a language agent's trajectory of  Think-Act-Observe by measuring the uncertainty of generated answers, and then choosing to accept the answer or to resort to external resources or human. This approach does not require any training and the language agent is prompted by few-shot learning. We find that this technique significantly improves the language agent’s performance on various QA tasks (regardless of the backbone LLM size), and offers a much more resource-efficient solution (substantially lower tool calling, and less number of output tokens). We also show that UALA outperforms fine-tuning language agents in the presence of small amount of data.

\section*{Limitations}
There are important limitations on UALA as presented. The approach presented here requires task-specific uncertainty selection and calibration set.  Additionally, we select the threshold based on the responses from Standard or CoT prompting of the LLM, which may not be the optimal choice for uncertainty within a trajectory. Moreover, we mainly focus on QA tasks and investigate limited types of tools. It is worth applying our framework on other types of tasks with different tools (e.g., mathematical reasoning tasks with program interpreters). Nonetheless, integrating uncertainty is a necessary element for designing intelligent language agents, and we believe that UALA can serve as the basis of more sophisticated techniques across domains.

\section*{Ethics Statement}
UALA is designed to serve as the basis of uncertainty-aware language agents and foster the research of creating more sophisticated language agents. This framework can also facilitate building uncertainty-aware AI agent applications and research based on open-source LLMs. Our framework does not involve training, hence we are not introducing new forms of biases through adjusting the models' weights. Nonetheless, UALA framework builds on existing LLMs and inherits their well-documented potential biases and of misuse associated with these models. Furthermore, as language agents like UALA utilise external tools, including search engines, they introduce an additional layer of potential bias and risk. The predictive behavior of the model is tailored by the outputs from these search engines. Our design allows for traceability of source (i.e., LLM vs. External engine), but we have not implemented any mitigation strategy for detecting these biases in the search engine responses. Similarly, our design does not have any mechanism to detect or prevent potential privacy violations that may arise through the search engine responses. 

\bibliography{anthology,custom}

\begin{thebibliography}{48}
\expandafter\ifx\csname natexlab\endcsname\relax\def\natexlab#1{#1}\fi

\bibitem[{Anil et~al.(2023)Anil, Dai, Firat, Johnson, Lepikhin, Passos, Shakeri, Taropa, Bailey, Chen, Chu, Clark, Shafey, Huang, Meier{-}Hellstern, Mishra, Moreira, Omernick, Robinson, Ruder, Tay, Xiao, Xu, Zhang, {\'{A}}brego, Ahn, Austin, Barham, Botha, Bradbury, Brahma, Brooks, Catasta, Cheng, Cherry, Choquette{-}Choo, Chowdhery, Crepy, Dave, Dehghani, Dev, Devlin, D{\'{\i}}az, Du, Dyer, Feinberg, Feng, Fienber, Freitag, Garcia, Gehrmann, Gonzalez, and et~al.}]{DBLP:journals/corr/abs-2305-10403}
Rohan Anil, Andrew~M. Dai, Orhan Firat, Melvin Johnson, Dmitry Lepikhin, Alexandre Passos, Siamak Shakeri, Emanuel Taropa, Paige Bailey, Zhifeng Chen, Eric Chu, Jonathan~H. Clark, Laurent~El Shafey, Yanping Huang, Kathy Meier{-}Hellstern, Gaurav Mishra, Erica Moreira, Mark Omernick, Kevin Robinson, Sebastian Ruder, Yi~Tay, Kefan Xiao, Yuanzhong Xu, Yujing Zhang, Gustavo~Hern{\'{a}}ndez {\'{A}}brego, Junwhan Ahn, Jacob Austin, Paul Barham, Jan~A. Botha, James Bradbury, Siddhartha Brahma, Kevin Brooks, Michele Catasta, Yong Cheng, Colin Cherry, Christopher~A. Choquette{-}Choo, Aakanksha Chowdhery, Cl{\'{e}}ment Crepy, Shachi Dave, Mostafa Dehghani, Sunipa Dev, Jacob Devlin, Mark D{\'{\i}}az, Nan Du, Ethan Dyer, Vladimir Feinberg, Fangxiaoyu Feng, Vlad Fienber, Markus Freitag, Xavier Garcia, Sebastian Gehrmann, Lucas Gonzalez, and et~al. 2023.
\newblock \href {https://doi.org/10.48550/ARXIV.2305.10403} {Palm 2 technical report}.
\newblock \emph{CoRR}, abs/2305.10403.

\bibitem[{Bacchus et~al.(1996)Bacchus, Grove, Halpern, and Koller}]{DBLP:journals/ai/BacchusGHK96}
Fahiem Bacchus, Adam~J. Grove, Joseph~Y. Halpern, and Daphne Koller. 1996.
\newblock \href {https://doi.org/10.1016/S0004-3702(96)00003-3} {From statistical knowledge bases to degrees of belief}.
\newblock \emph{Artif. Intell.}, 87(1-2):75--143.

\bibitem[{Brown et~al.(2020)Brown, Mann, Ryder, Subbiah, Kaplan, Dhariwal, Neelakantan, Shyam, Sastry, Askell, Agarwal, Herbert{-}Voss, Krueger, Henighan, Child, Ramesh, Ziegler, Wu, Winter, Hesse, Chen, Sigler, Litwin, Gray, Chess, Clark, Berner, McCandlish, Radford, Sutskever, and Amodei}]{DBLP:conf/nips/BrownMRSKDNSSAA20}
Tom~B. Brown, Benjamin Mann, Nick Ryder, Melanie Subbiah, Jared Kaplan, Prafulla Dhariwal, Arvind Neelakantan, Pranav Shyam, Girish Sastry, Amanda Askell, Sandhini Agarwal, Ariel Herbert{-}Voss, Gretchen Krueger, Tom Henighan, Rewon Child, Aditya Ramesh, Daniel~M. Ziegler, Jeffrey Wu, Clemens Winter, Christopher Hesse, Mark Chen, Eric Sigler, Mateusz Litwin, Scott Gray, Benjamin Chess, Jack Clark, Christopher Berner, Sam McCandlish, Alec Radford, Ilya Sutskever, and Dario Amodei. 2020.
\newblock \href {https://proceedings.neurips.cc/paper/2020/file/1457c0d6bfcb4967418bfb8ac142f64a-Paper.pdf} {Language models are few-shot learners}.
\newblock In \emph{Advances in Neural Information Processing Systems 33: Annual Conference on Neural Information Processing Systems 2020, NeurIPS 2020, December 6-12, 2020, virtual}.

\bibitem[{Chen et~al.(2023)Chen, Shu, Shareghi, Collier, Narasimhan, and Yao}]{DBLP:journals/corr/abs-2310-05915}
Baian Chen, Chang Shu, Ehsan Shareghi, Nigel Collier, Karthik Narasimhan, and Shunyu Yao. 2023.
\newblock \href {https://doi.org/10.48550/ARXIV.2310.05915} {Fireact: Toward language agent fine-tuning}.
\newblock \emph{CoRR}, abs/2310.05915.

\bibitem[{Chen et~al.(2022)Chen, Ma, Wang, and Cohen}]{DBLP:journals/corr/abs-2211-12588}
Wenhu Chen, Xueguang Ma, Xinyi Wang, and William~W. Cohen. 2022.
\newblock \href {https://doi.org/10.48550/ARXIV.2211.12588} {Program of thoughts prompting: Disentangling computation from reasoning for numerical reasoning tasks}.
\newblock \emph{CoRR}, abs/2211.12588.

\bibitem[{Collier et~al.(2022)Collier, Liu, and Shareghi}]{DBLP:journals/corr/abs-2208-11981}
Nigel~H. Collier, Fangyu Liu, and Ehsan Shareghi. 2022.
\newblock \href {https://doi.org/10.48550/ARXIV.2208.11981} {On reality and the limits of language data}.
\newblock \emph{CoRR}, abs/2208.11981.

\bibitem[{Fadeeva et~al.(2023)Fadeeva, Vashurin, Tsvigun, Vazhentsev, Petrakov, Fedyanin, Vasilev, Goncharova, Panchenko, Panov, Baldwin, and Shelmanov}]{fadeeva-etal-2023-lm}
Ekaterina Fadeeva, Roman Vashurin, Akim Tsvigun, Artem Vazhentsev, Sergey Petrakov, Kirill Fedyanin, Daniil Vasilev, Elizaveta Goncharova, Alexander Panchenko, Maxim Panov, Timothy Baldwin, and Artem Shelmanov. 2023.
\newblock \href {https://doi.org/10.18653/v1/2023.emnlp-demo.41} {{LM}-polygraph: Uncertainty estimation for language models}.
\newblock In \emph{Proceedings of the 2023 Conference on Empirical Methods in Natural Language Processing: System Demonstrations}, pages 446--461, Singapore. Association for Computational Linguistics.

\bibitem[{Gao et~al.(2023)Gao, Madaan, Zhou, Alon, Liu, Yang, Callan, and Neubig}]{DBLP:conf/icml/GaoMZ00YCN23}
Luyu Gao, Aman Madaan, Shuyan Zhou, Uri Alon, Pengfei Liu, Yiming Yang, Jamie Callan, and Graham Neubig. 2023.
\newblock \href {https://proceedings.mlr.press/v202/gao23f.html} {{PAL:} program-aided language models}.
\newblock In \emph{International Conference on Machine Learning, {ICML} 2023, 23-29 July 2023, Honolulu, Hawaii, {USA}}, volume 202 of \emph{Proceedings of Machine Learning Research}, pages 10764--10799. {PMLR}.

\bibitem[{Geva et~al.(2021)Geva, Khashabi, Segal, Khot, Roth, and Berant}]{DBLP:journals/tacl/GevaKSKRB21}
Mor Geva, Daniel Khashabi, Elad Segal, Tushar Khot, Dan Roth, and Jonathan Berant. 2021.
\newblock \href {https://doi.org/10.1162/TACL\_A\_00370} {Did aristotle use a laptop? {A} question answering benchmark with implicit reasoning strategies}.
\newblock \emph{Trans. Assoc. Comput. Linguistics}, 9:346--361.

\bibitem[{Gou et~al.(2023)Gou, Shao, Gong, Shen, Yang, Duan, and Chen}]{DBLP:journals/corr/abs-2305-11738}
Zhibin Gou, Zhihong Shao, Yeyun Gong, Yelong Shen, Yujiu Yang, Nan Duan, and Weizhu Chen. 2023.
\newblock \href {https://doi.org/10.48550/ARXIV.2305.11738} {{CRITIC:} large language models can self-correct with tool-interactive critiquing}.
\newblock \emph{CoRR}, abs/2305.11738.

\bibitem[{Hendrycks et~al.(2021)Hendrycks, Burns, Basart, Zou, Mazeika, Song, and Steinhardt}]{DBLP:conf/iclr/HendrycksBBZMSS21}
Dan Hendrycks, Collin Burns, Steven Basart, Andy Zou, Mantas Mazeika, Dawn Song, and Jacob Steinhardt. 2021.
\newblock \href {https://openreview.net/forum?id=d7KBjmI3GmQ} {Measuring massive multitask language understanding}.
\newblock In \emph{9th International Conference on Learning Representations, {ICLR} 2021, Virtual Event, Austria, May 3-7, 2021}. OpenReview.net.

\bibitem[{Hu et~al.(2022)Hu, yelong shen, Wallis, Allen-Zhu, Li, Wang, Wang, and Chen}]{hu2022lora}
Edward~J Hu, yelong shen, Phillip Wallis, Zeyuan Allen-Zhu, Yuanzhi Li, Shean Wang, Lu~Wang, and Weizhu Chen. 2022.
\newblock \href {https://openreview.net/forum?id=nZeVKeeFYf9} {Lo{RA}: Low-rank adaptation of large language models}.
\newblock In \emph{International Conference on Learning Representations}.

\bibitem[{Jiang et~al.(2023)Jiang, Xu, Gao, Sun, Liu, Dwivedi{-}Yu, Yang, Callan, and Neubig}]{DBLP:conf/emnlp/JiangXGSLDYCN23}
Zhengbao Jiang, Frank~F. Xu, Luyu Gao, Zhiqing Sun, Qian Liu, Jane Dwivedi{-}Yu, Yiming Yang, Jamie Callan, and Graham Neubig. 2023.
\newblock \href {https://doi.org/10.18653/V1/2023.EMNLP-MAIN.495} {Active retrieval augmented generation}.
\newblock In \emph{Proceedings of the 2023 Conference on Empirical Methods in Natural Language Processing, {EMNLP} 2023, Singapore, December 6-10, 2023}, pages 7969--7992. Association for Computational Linguistics.

\bibitem[{Kadavath et~al.(2022)Kadavath, Conerly, Askell, Henighan, Drain, Perez, Schiefer, Hatfield{-}Dodds, DasSarma, Tran{-}Johnson, Johnston, Showk, Jones, Elhage, Hume, Chen, Bai, Bowman, Fort, Ganguli, Hernandez, Jacobson, Kernion, Kravec, Lovitt, Ndousse, Olsson, Ringer, Amodei, Brown, Clark, Joseph, Mann, McCandlish, Olah, and Kaplan}]{DBLP:journals/corr/abs-2207-05221}
Saurav Kadavath, Tom Conerly, Amanda Askell, Tom Henighan, Dawn Drain, Ethan Perez, Nicholas Schiefer, Zac Hatfield{-}Dodds, Nova DasSarma, Eli Tran{-}Johnson, Scott Johnston, Sheer~El Showk, Andy Jones, Nelson Elhage, Tristan Hume, Anna Chen, Yuntao Bai, Sam Bowman, Stanislav Fort, Deep Ganguli, Danny Hernandez, Josh Jacobson, Jackson Kernion, Shauna Kravec, Liane Lovitt, Kamal Ndousse, Catherine Olsson, Sam Ringer, Dario Amodei, Tom Brown, Jack Clark, Nicholas Joseph, Ben Mann, Sam McCandlish, Chris Olah, and Jared Kaplan. 2022.
\newblock \href {https://doi.org/10.48550/ARXIV.2207.05221} {Language models (mostly) know what they know}.
\newblock \emph{CoRR}, abs/2207.05221.

\bibitem[{Kojima et~al.(2022)Kojima, Gu, Reid, Matsuo, and Iwasawa}]{kojima2022large}
Takeshi Kojima, Shixiang~Shane Gu, Machel Reid, Yutaka Matsuo, and Yusuke Iwasawa. 2022.
\newblock Large language models are zero-shot reasoners.
\newblock \emph{Advances in neural information processing systems}, 35:22199--22213.

\bibitem[{Kuhn et~al.(2023)Kuhn, Gal, and Farquhar}]{DBLP:conf/iclr/KuhnGF23}
Lorenz Kuhn, Yarin Gal, and Sebastian Farquhar. 2023.
\newblock \href {https://openreview.net/pdf?id=VD-AYtP0dve} {Semantic uncertainty: Linguistic invariances for uncertainty estimation in natural language generation}.
\newblock In \emph{The Eleventh International Conference on Learning Representations, {ICLR} 2023, Kigali, Rwanda, May 1-5, 2023}. OpenReview.net.

\bibitem[{Kumar and Byrne(2004)}]{kumar2004minimum}
Shankar Kumar and Bill Byrne. 2004.
\newblock Minimum bayes-risk decoding for statistical machine translation.
\newblock In \emph{Proceedings of the Human Language Technology Conference of the North American Chapter of the Association for Computational Linguistics: HLT-NAACL 2004}, pages 169--176.

\bibitem[{Lin et~al.(2023{\natexlab{a}})Lin, Fu, Yang, Ammanabrolu, Brahman, Huang, Bhagavatula, Choi, and Ren}]{DBLP:journals/corr/abs-2305-17390}
Bill~Yuchen Lin, Yicheng Fu, Karina Yang, Prithviraj Ammanabrolu, Faeze Brahman, Shiyu Huang, Chandra Bhagavatula, Yejin Choi, and Xiang Ren. 2023{\natexlab{a}}.
\newblock \href {https://doi.org/10.48550/ARXIV.2305.17390} {Swiftsage: {A} generative agent with fast and slow thinking for complex interactive tasks}.
\newblock \emph{CoRR}, abs/2305.17390.

\bibitem[{Lin et~al.(2022)Lin, Hilton, and Evans}]{DBLP:journals/tmlr/LinHE22}
Stephanie Lin, Jacob Hilton, and Owain Evans. 2022.
\newblock \href {https://openreview.net/forum?id=8s8K2UZGTZ} {Teaching models to express their uncertainty in words}.
\newblock \emph{Trans. Mach. Learn. Res.}, 2022.

\bibitem[{Lin et~al.(2023{\natexlab{b}})Lin, Trivedi, and Sun}]{DBLP:journals/corr/abs-2305-19187}
Zhen Lin, Shubhendu Trivedi, and Jimeng Sun. 2023{\natexlab{b}}.
\newblock \href {https://doi.org/10.48550/ARXIV.2305.19187} {Generating with confidence: Uncertainty quantification for black-box large language models}.
\newblock \emph{CoRR}, abs/2305.19187.

\bibitem[{Lu et~al.(2023)Lu, Peng, Cheng, Galley, Chang, Wu, Zhu, and Gao}]{lu2023chameleon}
Pan Lu, Baolin Peng, Hao Cheng, Michel Galley, Kai-Wei Chang, Ying~Nian Wu, Song-Chun Zhu, and Jianfeng Gao. 2023.
\newblock Chameleon: Plug-and-play compositional reasoning with large language models.
\newblock \emph{arXiv preprint arXiv:2304.09842}.

\bibitem[{Majumder et~al.(2023)Majumder, Mishra, Jansen, Tafjord, Tandon, Zhang, Callison{-}Burch, and Clark}]{DBLP:journals/corr/abs-2310-10134}
Bodhisattwa~Prasad Majumder, Bhavana~Dalvi Mishra, Peter~A. Jansen, Oyvind Tafjord, Niket Tandon, Li~Zhang, Chris Callison{-}Burch, and Peter Clark. 2023.
\newblock \href {https://doi.org/10.48550/ARXIV.2310.10134} {{CLIN:} {A} continually learning language agent for rapid task adaptation and generalization}.
\newblock \emph{CoRR}, abs/2310.10134.

\bibitem[{Malinin and Gales(2021)}]{DBLP:conf/iclr/MalininG21}
Andrey Malinin and Mark J.~F. Gales. 2021.
\newblock \href {https://openreview.net/forum?id=jN5y-zb5Q7m} {Uncertainty estimation in autoregressive structured prediction}.
\newblock In \emph{9th International Conference on Learning Representations, {ICLR} 2021, Virtual Event, Austria, May 3-7, 2021}. OpenReview.net.

\bibitem[{OpenAI(2023)}]{DBLP:journals/corr/abs-2303-08774}
OpenAI. 2023.
\newblock \href {https://doi.org/10.48550/ARXIV.2303.08774} {{GPT-4} technical report}.
\newblock \emph{CoRR}, abs/2303.08774.

\bibitem[{Ott et~al.(2018)Ott, Auli, Grangier, and Ranzato}]{DBLP:conf/icml/OttAGR18}
Myle Ott, Michael Auli, David Grangier, and Marc'Aurelio Ranzato. 2018.
\newblock \href {http://proceedings.mlr.press/v80/ott18a.html} {Analyzing uncertainty in neural machine translation}.
\newblock In \emph{Proceedings of the 35th International Conference on Machine Learning, {ICML} 2018, Stockholmsm{\"{a}}ssan, Stockholm, Sweden, July 10-15, 2018}, volume~80 of \emph{Proceedings of Machine Learning Research}, pages 3953--3962. {PMLR}.

\bibitem[{Patil et~al.(2023)Patil, Zhang, Wang, and Gonzalez}]{patil2023gorilla}
Shishir~G Patil, Tianjun Zhang, Xin Wang, and Joseph~E Gonzalez. 2023.
\newblock Gorilla: Large language model connected with massive apis.
\newblock \emph{arXiv preprint arXiv:2305.15334}.

\bibitem[{Petroni et~al.(2019)Petroni, Rockt{\"{a}}schel, Riedel, Lewis, Bakhtin, Wu, and Miller}]{DBLP:conf/emnlp/PetroniRRLBWM19}
Fabio Petroni, Tim Rockt{\"{a}}schel, Sebastian Riedel, Patrick S.~H. Lewis, Anton Bakhtin, Yuxiang Wu, and Alexander~H. Miller. 2019.
\newblock \href {https://doi.org/10.18653/V1/D19-1250} {Language models as knowledge bases?}
\newblock In \emph{Proceedings of the 2019 Conference on Empirical Methods in Natural Language Processing and the 9th International Joint Conference on Natural Language Processing, {EMNLP-IJCNLP} 2019, Hong Kong, China, November 3-7, 2019}, pages 2463--2473. Association for Computational Linguistics.

\bibitem[{Qiao et~al.(2023)Qiao, Gui, Chen, and Zhang}]{DBLP:journals/corr/abs-2305-13068}
Shuofei Qiao, Honghao Gui, Huajun Chen, and Ningyu Zhang. 2023.
\newblock \href {https://doi.org/10.48550/ARXIV.2305.13068} {Making language models better tool learners with execution feedback}.
\newblock \emph{CoRR}, abs/2305.13068.

\bibitem[{Ren et~al.(2023)Ren, Dixit, Bodrova, Singh, Tu, Brown, Xu, Takayama, Xia, Varley, Xu, Sadigh, Zeng, and Majumdar}]{DBLP:journals/corr/abs-2307-01928}
Allen~Z. Ren, Anushri Dixit, Alexandra Bodrova, Sumeet Singh, Stephen Tu, Noah Brown, Peng Xu, Leila Takayama, Fei Xia, Jake Varley, Zhenjia Xu, Dorsa Sadigh, Andy Zeng, and Anirudha Majumdar. 2023.
\newblock \href {https://doi.org/10.48550/ARXIV.2307.01928} {Robots that ask for help: Uncertainty alignment for large language model planners}.
\newblock \emph{CoRR}, abs/2307.01928.

\bibitem[{Schick et~al.(2023)Schick, Dwivedi{-}Yu, Dess{\`{\i}}, Raileanu, Lomeli, Zettlemoyer, Cancedda, and Scialom}]{DBLP:journals/corr/abs-2302-04761}
Timo Schick, Jane Dwivedi{-}Yu, Roberto Dess{\`{\i}}, Roberta Raileanu, Maria Lomeli, Luke Zettlemoyer, Nicola Cancedda, and Thomas Scialom. 2023.
\newblock \href {https://doi.org/10.48550/ARXIV.2302.04761} {Toolformer: Language models can teach themselves to use tools}.
\newblock \emph{CoRR}, abs/2302.04761.

\bibitem[{Shinn et~al.(2023)Shinn, Labash, and Gopinath}]{DBLP:journals/corr/abs-2303-11366}
Noah Shinn, Beck Labash, and Ashwin Gopinath. 2023.
\newblock \href {https://doi.org/10.48550/ARXIV.2303.11366} {Reflexion: an autonomous agent with dynamic memory and self-reflection}.
\newblock \emph{CoRR}, abs/2303.11366.

\bibitem[{Shu et~al.(2023)Shu, Han, Liu, Shareghi, and Collier}]{DBLP:conf/emnlp/ShuH0SC23}
Chang Shu, Jiuzhou Han, Fangyu Liu, Ehsan Shareghi, and Nigel Collier. 2023.
\newblock \href {https://aclanthology.org/2023.findings-emnlp.504} {{POSQA:} probe the world models of llms with size comparisons}.
\newblock In \emph{Findings of the Association for Computational Linguistics: {EMNLP} 2023, Singapore, December 6-10, 2023}, pages 7518--7531. Association for Computational Linguistics.

\bibitem[{Sumers et~al.(2023)Sumers, Yao, Narasimhan, and Griffiths}]{DBLP:journals/corr/abs-2309-02427}
Theodore~R. Sumers, Shunyu Yao, Karthik Narasimhan, and Thomas~L. Griffiths. 2023.
\newblock \href {https://doi.org/10.48550/ARXIV.2309.02427} {Cognitive architectures for language agents}.
\newblock \emph{CoRR}, abs/2309.02427.

\bibitem[{Suzgun et~al.(2023)Suzgun, Melas-Kyriazi, and Jurafsky}]{suzgun-etal-2023-follow}
Mirac Suzgun, Luke Melas-Kyriazi, and Dan Jurafsky. 2023.
\newblock \href {https://doi.org/10.18653/v1/2023.findings-acl.262} {Follow the wisdom of the crowd: Effective text generation via minimum {B}ayes risk decoding}.
\newblock In \emph{Findings of the Association for Computational Linguistics: ACL 2023}, pages 4265--4293, Toronto, Canada. Association for Computational Linguistics.

\bibitem[{Tian et~al.(2023)Tian, Mitchell, Zhou, Sharma, Rafailov, Yao, Finn, and Manning}]{DBLP:conf/emnlp/TianMZSRYFM23}
Katherine Tian, Eric Mitchell, Allan Zhou, Archit Sharma, Rafael Rafailov, Huaxiu Yao, Chelsea Finn, and Christopher~D. Manning. 2023.
\newblock \href {https://aclanthology.org/2023.emnlp-main.330} {Just ask for calibration: Strategies for eliciting calibrated confidence scores from language models fine-tuned with human feedback}.
\newblock In \emph{Proceedings of the 2023 Conference on Empirical Methods in Natural Language Processing, {EMNLP} 2023, Singapore, December 6-10, 2023}, pages 5433--5442. Association for Computational Linguistics.

\bibitem[{Touvron et~al.(2023)Touvron, Martin, Stone, Albert, Almahairi, Babaei, Bashlykov, Batra, Bhargava, Bhosale, Bikel, Blecher, Canton{-}Ferrer, Chen, Cucurull, Esiobu, Fernandes, Fu, Fu, Fuller, Gao, Goswami, Goyal, Hartshorn, Hosseini, Hou, Inan, Kardas, Kerkez, Khabsa, Kloumann, Korenev, Koura, Lachaux, Lavril, Lee, Liskovich, Lu, Mao, Martinet, Mihaylov, Mishra, Molybog, Nie, Poulton, Reizenstein, Rungta, Saladi, Schelten, Silva, Smith, Subramanian, Tan, Tang, Taylor, Williams, Kuan, Xu, Yan, Zarov, Zhang, Fan, Kambadur, Narang, Rodriguez, Stojnic, Edunov, and Scialom}]{DBLP:journals/corr/abs-2307-09288}
Hugo Touvron, Louis Martin, Kevin Stone, Peter Albert, Amjad Almahairi, Yasmine Babaei, Nikolay Bashlykov, Soumya Batra, Prajjwal Bhargava, Shruti Bhosale, Dan Bikel, Lukas Blecher, Cristian Canton{-}Ferrer, Moya Chen, Guillem Cucurull, David Esiobu, Jude Fernandes, Jeremy Fu, Wenyin Fu, Brian Fuller, Cynthia Gao, Vedanuj Goswami, Naman Goyal, Anthony Hartshorn, Saghar Hosseini, Rui Hou, Hakan Inan, Marcin Kardas, Viktor Kerkez, Madian Khabsa, Isabel Kloumann, Artem Korenev, Punit~Singh Koura, Marie{-}Anne Lachaux, Thibaut Lavril, Jenya Lee, Diana Liskovich, Yinghai Lu, Yuning Mao, Xavier Martinet, Todor Mihaylov, Pushkar Mishra, Igor Molybog, Yixin Nie, Andrew Poulton, Jeremy Reizenstein, Rashi Rungta, Kalyan Saladi, Alan Schelten, Ruan Silva, Eric~Michael Smith, Ranjan Subramanian, Xiaoqing~Ellen Tan, Binh Tang, Ross Taylor, Adina Williams, Jian~Xiang Kuan, Puxin Xu, Zheng Yan, Iliyan Zarov, Yuchen Zhang, Angela Fan, Melanie Kambadur, Sharan Narang, Aur{\'{e}}lien Rodriguez, Robert Stojnic, Sergey Edunov,
  and Thomas Scialom. 2023.
\newblock \href {https://doi.org/10.48550/arXiv.2307.09288} {Llama 2: Open foundation and fine-tuned chat models}.
\newblock \emph{CoRR}, abs/2307.09288.

\bibitem[{Wang et~al.(2023)Wang, Wei, Schuurmans, Le, Chi, Narang, Chowdhery, and Zhou}]{DBLP:conf/iclr/0002WSLCNCZ23}
Xuezhi Wang, Jason Wei, Dale Schuurmans, Quoc~V. Le, Ed~H. Chi, Sharan Narang, Aakanksha Chowdhery, and Denny Zhou. 2023.
\newblock \href {https://openreview.net/pdf?id=1PL1NIMMrw} {Self-consistency improves chain of thought reasoning in language models}.
\newblock In \emph{The Eleventh International Conference on Learning Representations, {ICLR} 2023, Kigali, Rwanda, May 1-5, 2023}. OpenReview.net.

\bibitem[{Wei et~al.(2022)Wei, Wang, Schuurmans, Bosma, Ichter, Xia, Chi, Le, and Zhou}]{DBLP:conf/nips/Wei0SBIXCLZ22}
Jason Wei, Xuezhi Wang, Dale Schuurmans, Maarten Bosma, Brian Ichter, Fei Xia, Ed~H. Chi, Quoc~V. Le, and Denny Zhou. 2022.
\newblock \href {https://openreview.net/pdf?id=_VjQlMeSB_J} {Chain-of-thought prompting elicits reasoning in large language models}.
\newblock In \emph{NeurIPS}.

\bibitem[{Wiher et~al.(2022)Wiher, Meister, and Cotterell}]{wiher-etal-2022-decoding}
Gian Wiher, Clara Meister, and Ryan Cotterell. 2022.
\newblock \href {https://doi.org/10.1162/tacl_a_00502} {On decoding strategies for neural text generators}.
\newblock \emph{Transactions of the Association for Computational Linguistics}, 10:997--1012.

\bibitem[{Xi et~al.(2023)Xi, Chen, Guo, He, Ding, Hong, Zhang, Wang, Jin, Zhou, Zheng, Fan, Wang, Xiong, Zhou, Wang, Jiang, Zou, Liu, Yin, Dou, Weng, Cheng, Zhang, Qin, Zheng, Qiu, Huan, and Gui}]{DBLP:journals/corr/abs-2309-07864}
Zhiheng Xi, Wenxiang Chen, Xin Guo, Wei He, Yiwen Ding, Boyang Hong, Ming Zhang, Junzhe Wang, Senjie Jin, Enyu Zhou, Rui Zheng, Xiaoran Fan, Xiao Wang, Limao Xiong, Yuhao Zhou, Weiran Wang, Changhao Jiang, Yicheng Zou, Xiangyang Liu, Zhangyue Yin, Shihan Dou, Rongxiang Weng, Wensen Cheng, Qi~Zhang, Wenjuan Qin, Yongyan Zheng, Xipeng Qiu, Xuanjing Huan, and Tao Gui. 2023.
\newblock \href {https://doi.org/10.48550/ARXIV.2309.07864} {The rise and potential of large language model based agents: {A} survey}.
\newblock \emph{CoRR}, abs/2309.07864.

\bibitem[{Xiao and Wang(2021)}]{DBLP:conf/eacl/XiaoW21}
Yijun Xiao and William~Yang Wang. 2021.
\newblock \href {https://doi.org/10.18653/V1/2021.EACL-MAIN.236} {On hallucination and predictive uncertainty in conditional language generation}.
\newblock In \emph{Proceedings of the 16th Conference of the European Chapter of the Association for Computational Linguistics: Main Volume, {EACL} 2021, Online, April 19 - 23, 2021}, pages 2734--2744. Association for Computational Linguistics.

\bibitem[{Xu et~al.(2020)Xu, Desai, and Durrett}]{xu-etal-2020-understanding-neural}
Jiacheng Xu, Shrey Desai, and Greg Durrett. 2020.
\newblock \href {https://doi.org/10.18653/v1/2020.emnlp-main.508} {Understanding neural abstractive summarization models via uncertainty}.
\newblock In \emph{Proceedings of the 2020 Conference on Empirical Methods in Natural Language Processing (EMNLP)}, pages 6275--6281, Online. Association for Computational Linguistics.

\bibitem[{Yang et~al.(2023)Yang, Li, Wang, and Wang}]{DBLP:journals/corr/abs-2310-04782}
Yuchen Yang, Houqiang Li, Yanfeng Wang, and Yu~Wang. 2023.
\newblock \href {https://doi.org/10.48550/ARXIV.2310.04782} {Improving the reliability of large language models by leveraging uncertainty-aware in-context learning}.
\newblock \emph{CoRR}, abs/2310.04782.

\bibitem[{Yang et~al.(2018)Yang, Qi, Zhang, Bengio, Cohen, Salakhutdinov, and Manning}]{DBLP:conf/emnlp/Yang0ZBCSM18}
Zhilin Yang, Peng Qi, Saizheng Zhang, Yoshua Bengio, William~W. Cohen, Ruslan Salakhutdinov, and Christopher~D. Manning. 2018.
\newblock \href {https://doi.org/10.18653/V1/D18-1259} {Hotpotqa: {A} dataset for diverse, explainable multi-hop question answering}.
\newblock In \emph{Proceedings of the 2018 Conference on Empirical Methods in Natural Language Processing, Brussels, Belgium, October 31 - November 4, 2018}, pages 2369--2380. Association for Computational Linguistics.

\bibitem[{Yao et~al.(2023)Yao, Zhao, Yu, Du, Shafran, Narasimhan, and Cao}]{DBLP:conf/iclr/YaoZYDSN023}
Shunyu Yao, Jeffrey Zhao, Dian Yu, Nan Du, Izhak Shafran, Karthik~R. Narasimhan, and Yuan Cao. 2023.
\newblock \href {https://openreview.net/pdf?id=WE\_vluYUL-X} {React: Synergizing reasoning and acting in language models}.
\newblock In \emph{The Eleventh International Conference on Learning Representations, {ICLR} 2023, Kigali, Rwanda, May 1-5, 2023}. OpenReview.net.

\bibitem[{Yin et~al.(2023)Yin, Brahman, Ravichander, Chandu, Chang, Choi, and Lin}]{yin2023lumos}
Da~Yin, Faeze Brahman, Abhilasha Ravichander, Khyathi Chandu, Kai-Wei Chang, Yejin Choi, and Bill~Yuchen Lin. 2023.
\newblock Lumos: Learning agents with unified data, modular design, and open-source llms.
\newblock \emph{arXiv preprint arXiv:2311.05657}.

\bibitem[{Zablotskaia et~al.(2023)Zablotskaia, Phan, Maynez, Narayan, Ren, and Liu}]{DBLP:conf/emnlp/ZablotskaiaPMN023}
Polina Zablotskaia, Du~Phan, Joshua Maynez, Shashi Narayan, Jie Ren, and Jeremiah~Z. Liu. 2023.
\newblock \href {https://aclanthology.org/2023.findings-emnlp.197} {On uncertainty calibration and selective generation in probabilistic neural summarization: {A} benchmark study}.
\newblock In \emph{Findings of the Association for Computational Linguistics: {EMNLP} 2023, Singapore, December 6-10, 2023}, pages 2980--2992. Association for Computational Linguistics.

\bibitem[{Zhou et~al.(2023)Zhou, Jiang, Li, Wu, Wang, Qiu, Zhang, Chen, Wu, Wang, Zhu, Chen, Zhang, Zhang, Chen, Cui, and Sachan}]{DBLP:journals/corr/abs-2309-07870}
Wangchunshu Zhou, Yuchen~Eleanor Jiang, Long Li, Jialong Wu, Tiannan Wang, Shi Qiu, Jintian Zhang, Jing Chen, Ruipu Wu, Shuai Wang, Shiding Zhu, Jiyu Chen, Wentao Zhang, Ningyu Zhang, Huajun Chen, Peng Cui, and Mrinmaya Sachan. 2023.
\newblock \href {https://doi.org/10.48550/ARXIV.2309.07870} {Agents: An open-source framework for autonomous language agents}.
\newblock \emph{CoRR}, abs/2309.07870.

\end{thebibliography}
\bibliographystyle{acl_natbib}

\appendix
\section*{Appendix}
\section{Effect of Quantile Value}
\label{sec:quantile_vale}
\begin{figure}[t]
\centering
  \includegraphics[trim={0.2cm, 0.3cm, 10.3cm, 0.2cm}, clip, scale=0.5]{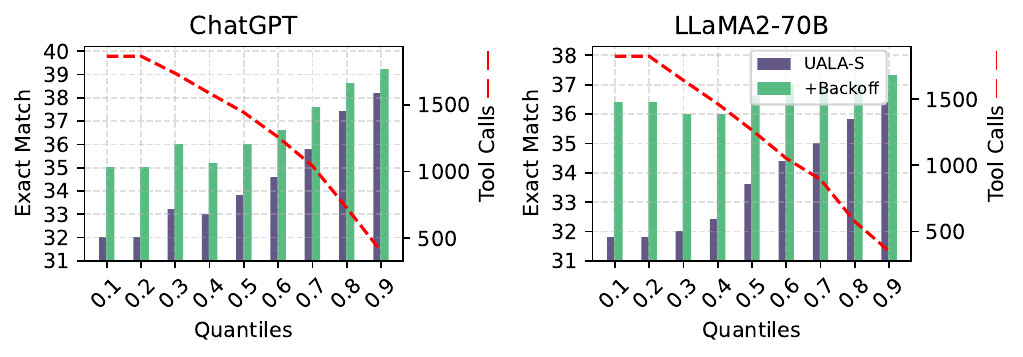}
  \includegraphics[trim={9.8cm, 0.3cm, 0.3cm, 0.2cm}, clip, scale=0.5]{figures/quantiles.pdf}
  \label{fig:kl}
\caption{The results on HotpotQA, using different quantile values for uncertainty threshold estimation in single-inference with ChatGPT (left) and LLaMA2-70B (right). The uncertainty estimation method used is Entropy.}
\label{fig:quantile_value}
\end{figure}

To investigate the effect of the quantile value used in single-inference uncertainty threshold estimation, we compare the results of using different quantile values from 0.1 to 0.9 on HotpotQA using ChatGPT and LLaMA2-70B, shown in Figure~\ref{fig:quantile_value}. As the increase of quantile value, the number of instances with tool-use decreases. This is because the quantile is based on the uncertainty of the calibration set, larger quantile value indicates higher uncertainty threshold, which leads to less tool-use. For UALA-S, large quantile value (i.e., 0.9) has the best performance on both LLMs. When using backoff, the trend is the same on ChatGPT, but LLaMA2-70B has the opposite trend. Based on our observation, we find the reason is that LLaMA2-70B is more likely to generate a null answer than ChatGPT (generate a wrong answer) when the external tool fails to provide useful information, which can be amended by the backoff setting. However, UALA-S+Backoff on ChatGPT with a large quantile value can achieve the same good performance but with only a small number of tool-use.

\section{Integrating UALA in the CRITIC Framework}
\label{sec:critic}
The CRITIC~\cite{DBLP:journals/corr/abs-2305-11738} enables the LLMs to verify and iteratively self-correct their output through interaction with external tools. Following the CRITIC~\cite{DBLP:journals/corr/abs-2305-11738} work, we also compare two settings: using Google Search as the external tool and using the LLM instead of an external tool (CRITIC w/o Tool) without changing the prompt of CRITIC to generate evidence. Based on CoT outputs, we conduct standard CRITIC and our uncertainty-based CRITIC. Specifically, based on the uncertainty of the answer, we choose whether to do CRITIC or keep the answer. For single-inference estimation, we use Log-sum and 0.8 quantile for threshold estimation. The multi-inference uncertainty estimation is based on 9 samples generated using temperature $0.7$. 

\begin{table}[t]
\centering
\resizebox{0.85\columnwidth}{!}{%
\begin{tabular}{lc}
\toprule
Methods              & EM        \\ \hline
CoT                  & 34.8(0)   \\
CRITIC               & 41.0(1,500) \\
CRITIC w/o Tool      & 35.6(1,500) \\ \hline
UALA-S-CRITIC          & 39.0(597) \\
UALA-S-CRITIC w/o Tool & 38.0(597) \\
UALA-M-CRITIC          & 40.6(795) \\
UALA-M-CRITIC w/o Tool & 37.4(795) \\ \bottomrule
\end{tabular}}
\caption{HotpotQA Results using CRITIC~\cite{DBLP:journals/corr/abs-2305-11738} on ChatGPT. The round of iterative correction in CRITIC is three.}
\label{tab:critic}
\end{table}

The results are demonstrated in Table~\ref{tab:critic}. Standard CRITIC performs much better than CRITIC w/o Tool and both of them improve the CoT result effectively. Our uncertainty-based methods reduce the frequency of tool-use  by nearly half, but still achieve great results on par with the standard CRITIC. Moreover, both of the S-Un-CRITIC w/o Tool and M-Un-CRITIC w/o Tool achieve better results than standard CRITIC w/o Tool. These results indicate the effectiveness and generalisation of our proposed uncertainty-based tool-use methods with regard to different tool-use frameworks.

\begin{table}[t]
\centering
\resizebox{0.9\columnwidth}{!}{%
\begin{tabular}{lccccc}
\hline
\multicolumn{1}{c}{\multirow{2}{*}{\textbf{Models}}} & \multicolumn{5}{c} 
{\textbf{Methods}}                \\
\multicolumn{1}{c}{}                                 & Standard & CoT   & ReAct  & UALA-S & UALA-M \\ \hline
\multicolumn{6}{c}{HotpotQA}                                                                               \\
ChatGPT                                              & 3,261     & 29,152 & 308,480 & 86,893      & 460,366     \\
LLaMA2-70B                                           & 1,997     & 28,957 & 403,132 & 95,984      & 496,604     \\ \hline
\multicolumn{6}{c}{StrategyQA}                                                                             \\
ChatGPT                                              & 1,134     & 23,222 & 113,106 & 22,721      & 48,709      \\
LLaMA2-70B                                           & 229      & 23,999 & 169,543 & 56,916      & 95,471      \\ \hline
\multicolumn{6}{c}{MMLU}                                                                                   \\
ChatGPT                                              & 574      & 61,202 & 200,450 & 75,095      & 78,453      \\
LLaMA2-70B                                           & 570      & 55,761 & 320,151 & 159,367     & 144,589    \\ \hline
\end{tabular}%
}
\caption{The total number of output tokens for each method on ChatGPT and LLaMA2-70B.}
\vspace{-4mm}
\label{tab:num_tokenx}
\end{table}

\section{Uncertainty-aware Method (UALA) vs. Fine-tuning Method (FireAct)}
Based on the published results of ~\citet{DBLP:journals/corr/abs-2310-05915}, we also compare results between the uncertainty-aware method (UALA) and fine-tuning method (FireAct) in Table~\ref{tab:fireact_vs_uala}. FireAct uses a mix of training data from these three datasets (2,470 samples in total) to do multi-task learning (the setting reported as their best result). We report the results provided in their paper and calculate its improvement of over ReAct. For the UALA, we calculate the improvement of UALA-M+Backoff setting over ReAct. Our UALA obtains more improvements than FireAct over all three tasks. UALA also requires less training data to construct the calibration set compared with fine-tuning data creation in FireAct. For instance, on HotpotQA, FireAct collects 2,000 training questions for fine-tuning data curation, while UALA only uses 500 training questions to create calibration set.

\begin{table*}[!htbp]
\centering
\begin{tabular}{lcccccc}
\hline
\multicolumn{1}{c}{\multirow{2}{*}{\textbf{Tasks}}} & \multicolumn{3}{c}{\textbf{Fine-tuning Improvement}} & \multicolumn{3}{c}{\textbf{Uncertainty-aware Improvement}} \\
\multicolumn{1}{c}{}                                & ReAct        & FireAct        & abs./rel. diff       & ReAct            & UALA            & abs./rel. diff            \\ \cmidrule(lr){2-4}\cmidrule(lr){5-7}
HotpotQA                                            & 31.4         & 39.6           & ↑8.2/26.1\%          & 32.0             & 41.2            & ↑9.2/28.8\%               \\
StrategyQA                                          & 61.0         & 72.9           & ↑11.9/19.5\%         & 55.5             & 66.9            & ↑11.4/20.5\%              \\
MMLU                                                & 58.6         & 65.8           & ↑7.2/12.3\%          & 55.8             & 72.2            & ↑16.4/29.3\%   \\ \hline
\end{tabular}
\caption{The comparison between fine-tuning method (FireAct) and our uncertainty-aware method (UALA). Since the data split is not exactly the same for all tasks, we report the comparison results with ReAct. FireAct uses multi-task learning results (numbers provided in their paper) and UALA uses UALA-M+Backoff results. Both methods are on ChatGPT (fine-tune vs. few-shot learning).}
\label{tab:fireact_vs_uala}
\end{table*}

\section{Effect of Different Sizes of Calibration Set for UALA-S}
\label{subsec:calib-size}
The single-inference uncertainty threshold estimation is based on a calibration set which consists of correctly answered questions (Section~\ref{sec:sin-inf-thre}). To probe the effect of calibration set's size, we vary the set size (200 to 2000 instances) using the same uncertainty estimation setting. The performance has a negligible fluctuation between 37.6-38.2, while the Pearson correlation coefficient of 0.29 and the p-value of 0.39 ($p < .05$) indicate no statistically significant correlation between the calibration set size and the downstream performance. This suggests a calibration set of less than 200 examples could suffice in practice. See Table~\ref{tab:calibration_size_plot} for details.

\begin{table*}[!htbp]
    \centering

    \scalebox{0.6}{
    \begin{tabular}{lccccccccccc}
    \toprule
        Calibration Set Size &178&386&558&745&937&1132&1330&1546&1752&1960&2145\\
Performance&38.2(403)&37.6(503)&37.6(503)&38.2(485)&38.2(485)&37.6(503)&37.6(503)&37.8(500)&38.2(485)&38.2(485)&38.2(485)\\
\bottomrule 
    \end{tabular}}
        \caption{Results of different sizes of calibration set for single-inference uncertainty threshold estimation (UALA-S) on HotpotQA using ChatGPT. Numbers in parenthesis are tool calls.}
    \label{tab:calibration_size_plot}
\end{table*}

\section{Single-inference Estimation Methods Comparison Results}
\label{sec:sin-es-me}
For single-inference estimation of the free-form answer, we investigate the effect of five methods described in Section~\ref{sec:free-form} and the uncertainty threshold estimation methods described in Section~\ref{sec:sin-inf-thre}. The results (on 500 instances of HotpotQA dev set) are shown in Table~\ref{tab: diff_settings}. Most settings perform better than CoT and ReAct. When using Max and Quantile as the threshold estimation method, there is no significant difference between these five uncertainty estimation methods on both LLMs. For the Mean threshold estimation method, Normalised Product and Log-sum perform better than other three uncertainty estimation methods. It also has more tool-use instances than the other two threshold estimation methods. Overall, the Quantile achieves the best performance on both LLMs.

\begin{table*}[!htbp]\footnotesize
\centering
\resizebox{\textwidth}{!}{%
\begin{tabular}{lcccccc}
\toprule
                                      & \multicolumn{6}{c}{\textbf{Uncertainty Threshold Estimation Methods}}                                           \\
                                      & \multicolumn{3}{c}{\textbf{ChatGPT}}             & \multicolumn{3}{c}{\textbf{LLaMA2-70B}}          \\
                                      \cmidrule(lr){2-4}\cmidrule(lr){5-7}
 \textbf{Uncertainty Estimation Methods}  & Max & Mean & Quantile & Max & Mean & Quantile \\ 
Minimum                                 & 35.6(90)     & 33.0(1,531)     & 38.0(486)         & 35.8(80)     & 32.2(1,567)     & 35.8(352)          \\
Average                                 & 35.8(110)     & 33.0(1,531)     & 38.2(406)          & 35.8(68)     & 32.2(1,567)     & 36.2(380)          \\
Normalised Product                      & 35.8(121)     & 36.8(943)     & 38.2(478)         & 35.8(77)     & 34.2(1,196)     & 35.8(352)          \\
Log-sum                                 & 35.8(121)     & 36.8(943)     & 38.2(478)         & 35.8(77)     & 34.2(1,196)     & 35.8(352)          \\
Entropy                                 & 35.8(114)     & 33.0(1,531)     & \textbf{38.2(403)} & 35.6(51)     & 32.2(1,567)     & \textbf{36.4(350)} \\ \bottomrule
\end{tabular}}
\caption{Results on HotpotQA comparing different settings of single-inference uncertainty estimation methods (five methods) and uncertainty threshold estimation methods (three methods) on ChatGPT and LLaMA2-70B. We use 0.9 as quantile value for Quantile method. \textbf{Bold} shows the best result for each LLM among all different settings. The CoT results with no tool calls are (ChatGPT, EM: 34.8) and (LLaMA2-70B, EM: 35.6). The ReAct results with 100\% tool calls are (ChatGPT, EM: 32.0) and (LLaMA2-70B, EM: 32.4).}
\label{tab: diff_settings}
\end{table*}

\section{Comparison of Uncertainty for Correct vs. Incorrect Answers}
As described in Section~\ref{sec:corr_vs_incorr_uncertainty}, we  calculated the mean uncertainty of correct and incorrect answers. Table~\ref{table:stat_comparison} demonstrates the difference of the two value across three datasets, and various other statistical tests on the means to highlight the significance of the difference between uncertainties for correct and incorrect responses.

\begin{table*}[!htbp]
\centering
\scalebox{0.85}{
\begin{tabular}{llccccc}
\toprule
&\textbf{Dataset} & \textbf{Sample Size} & \textbf{Difference in Means} & \textbf{T-test} & \textbf{P-value} & \textbf{Cohen's d} \\ \hline
\multirow{3}{*}{Single-Inference}&HotpotQA         & 500                  & 0.68                         & 3.0382              & 0.0039758        & 0.7845             \\ 
&StrategyQA       & 229                  & 0.40                         & 4.2226              & 0.00012028       & 1.0903             \\ 
&MMLU             & 570                  & 0.34                         & 4.6761              & 0.00001912       & 1.2074             \\ \hline
\multirow{3}{*}{Multiple-Inference}&HotpotQA         & 500                  & 0.44                         & 24.2285              & \(3.34 \times 10^{-101}\) & 1.5323\\ 
&StrategyQA       & 229                  & 0.32                         & 9.4750               & \(3.27 \times 10^{-19}\)  & 0.8855\\ 
&MMLU             & 570                  &0.21                         & 16.5838              & \(9.11 \times 10^{-55}\)  & 0.9823            \\ \bottomrule
\end{tabular}}
\caption{Comparison of uncertainty for Correct vs. Incorrect answers on test set based on mean and standard deviation: Means Difference, T-test, Cohen's effect size on HotpotQA, StrategyQA, and MMLU. The backbone LLM is ChatGPT.}
\label{table:stat_comparison}
\end{table*}


\begin{table*}[!htbp]
\centering
\resizebox{\textwidth}{!}{%
\begin{tabular}{lcccccccc}
\toprule
\multicolumn{1}{c}{\multirow{2}{*}{\textbf{Methods}}} & \multicolumn{4}{c}{\textbf{ChatGPT}}                                                         & \multicolumn{4}{c}{\textbf{LLaMA2-70B}}                                                      \\ \cmidrule(lr){2-5}\cmidrule(lr){6-9}
\multicolumn{1}{c}{}                                  & \textbf{HotpotQA} & \textbf{StrategyQA} & \textbf{MMLU}  & \multicolumn{1}{c}{\textbf{Avg.}} & \textbf{HotpotQA} & \textbf{StrategyQA} & \textbf{MMLU}  & \multicolumn{1}{c}{\textbf{Avg.}} \\ \hline
                                                      & \multicolumn{8}{c}{Absolute (Relative) Performance Improvement}                                                                                                                    \\ 
UALA-S+Backoff                                        & ↑4.4(12.6\%)      & ↑8.8(15.2\%)       & ↑2.4(3.5\%)    & ↑5.6(10.4\%)                      & ↑1.7(4.8\%)       & ↑5.7(8.6\%)        & ↑5.1(7.9\%)  & ↑4.2(7.1\%)                      \\
UALA-M+Backoff                                        & ↑6.5(18.7\%)     & ↑9.3(16.1\%)        & ↑3.2(4.6\%)  & ↑6.3(13.1\%)                     & ↑4.6(12.9\%)      & ↑5.9(9.0\%)        & ↑5.6(8.7\%)  & ↑5.4(10.2\%)                     \\ \hline
                                                      & \multicolumn{8}{c}{Absolute (Relative) Tool Calls Decrement}                                                                                                                       \\ 
UALA-S+Backoff                                        & ↓1,711(80.9\%)    & ↓575(81.1\%)        & ↓1,162(63.7\%) & ↓1,149(75.2\%)                    & ↓1,744(83.3\%)    & ↓592(66.5\%)        & ↓1,454(51.8\%) & ↓1,263(67.2\%)                    \\
UALA-M+Backoff                                        & ↓931(44.0\%)      & ↓475(67.0\%)        & ↓1,183(64.9\%)  & ↓863(58.6\%)                      & ↓1,169(55.8\%)      & ↓318(35.7\%)        & ↓1,612(57.4\%) & ↓1,033(49.6\%)       \\ \bottomrule              
\end{tabular}%
}
\caption{The statistics of absolute (relative) performance improvement between UALA+Backoff methods and Baselines (CoT for HotpotQA, and Standard for StrategyQA and MMLU); absolute (relative) tool calls decrement between UALA+Backoff and ReAct.}
\label{tab:result_statistic}
\end{table*}

\section{Verbalised Confidence Prompt Template}
\label{sec:verbalised_template}
\texttt{Solve a question answering task. Your task is to generate Thought and Answer where a Thought can reason about the current situation by thinking step by step. Provide the probability that the answer is correct (0.0 to 1.0). Give ONLY the probability in the format of Answer[Probability], no other words or explanation.}

\pagebreak
\onecolumn
\section{Full Prompt Templates}
\label{appd:prompt_template}
\begin{xltabular}{\textwidth}{p{\textwidth}}
\toprule
\multicolumn{1}{c}{\textbf{HotpotQA}}                                                                  \\ \hline
\textbf{Standard} \\
Answer the question:\\
Question: What is the elevation range for the area that the eastern sector of the Colorado orogeny extends into?\\ Answer: 1,800 to 7,000 ft \\ \\
Question: Musician and satirist Allie Goertz wrote a song about the "The Simpsons" character Milhouse, who Matt Groening named after who?\\ Answer: Richard Nixon \\ \\
Question: Which documentary is about Finnish rock groups, Adam Clayton Powell or The Saimaa Gesture?\\ Answer: The Saimaa Gesture \\ \\
Question: What profession does Nicholas Ray and Elia Kazan have in common? \\ Answer: director, screenwriter, actor \\ \\
Question: Which magazine was started first Arthur's Magazine or First for Women? \\ Answer: Arthur's Magazine \\ \\
Question: Were Pavel Urysohn and Leonid Levin known for the same type of work? \\ Answer: Yes \\ \hline
\textbf{Chain-of-Thought (CoT)} \\ 
Solve a question answering task. Your task is to generate Thought and Answer where a Thought can reason about the current situation by thinking step by step. \\
Here are some examples. \\
Question: What is the elevation range for the area that the eastern sector of the Colorado orogeny extends into?\\ Thought: Let's think step by step. The eastern sector of Colorado orogeny extends into the High Plains. High Plains rise in elevation from around 1,800 to 7,000 ft, so the answer is 1,800 to 7,000 ft. \\ Answer: 1,800 to 7,000 ft \\ \\
Question: Musician and satirist Allie Goertz wrote a song about the "The Simpsons" character Milhouse, who Matt Groening named after who? \\ Thought: Let's think step by step. Milhouse was named after U.S. president Richard Nixon, so the answer is Richard Nixon. \\ Answer: Richard Nixon \\ \\
Question: Which documentary is about Finnish rock groups, Adam Clayton Powell or The Saimaa Gesture?\\ Thought: Let's think step by step. Adam Clayton Powell (film) is a documentary about an African-American politician, not Finnish rock groups. So the documentary about Finnish rock groups must instead be The Saimaa Gesture. \\ Answer: The Saimaa Gesture \\ \\
Question: What profession does Nicholas Ray and Elia Kazan have in common? \\ Thought: Let's think step by step. Professions of Nicholas Ray are director, screenwriter, and actor. Professions of Elia Kazan are director, producer, screenwriter, and actor. So profession Nicholas Ray and Elia Kazan have in common is director, screenwriter, and actor.\\ Answer: director, screenwriter, actor \\ \\
Question: Which magazine was started first Arthur's Magazine or First for Women? \\ Thought: Let's think step by step. Arthur's Magazine was started in 1844. First for Women was started in 1989. 1844 (Arthur's Magazine) $<$ 1989 (First for Women), so Arthur's Magazine was started first. \\ Answer: Arthur's Magazine \\ \\
Question: Were Pavel Urysohn and Leonid Levin known for the same type of work? \\ Thought: Let's think step by step. Pavel Urysohn is a mathematician. Leonid Levin is a mathematician and computer scientist. So Pavel Urysohn and Leonid Levin have the same type of work. \\ Answer: Yes \\ \hline
\textbf{ReAct} \\ 
Solve a question answering task with interleaving Thought, Action, Observation steps. Thought can reason about the current situation, and Action can be three types: \\
(1) Search[entity], which searches the exact entity on Wikipedia and returns the first paragraph if it exists. If not, it will return some similar entities to search. \\ 
(2) Lookup[keyword], which returns the next sentence containing keyword in the current passage.
(3) Finish[answer], which returns the answer and finishes the task. \\
Here are some examples. \\
Question: What is the elevation range for the area that the eastern sector of the Colorado orogeny extends into?\\ Thought 1: I need to search Colorado orogeny, find the area that the eastern sector of the Colorado orogeny extends into, then find the elevation range of the area.\\ Action 1: Search[Colorado orogeny]\\Observation 1: The Colorado orogeny was an episode of mountain building (an orogeny) in Colorado and surrounding areas.\\Thought 2: It does not mention the eastern sector. So I need to look up eastern sector.\\Action 2: Lookup[eastern sector]\\Observation 2: (Result 1 / 1) The eastern sector extends into the High Plains and is called the Central Plains orogeny.\\Thought 3: The eastern sector of Colorado orogeny extends into the High Plains. So I need to search High Plains and find its elevation range.\\Action 3: Search[High Plains]\\Observation 3: High Plains refers to one of two distinct land regions:\\Thought 4: I need to instead search High Plains (United States).\\Action 4: Search[High Plains (United States)]\\Observation 4: The High Plains are a subregion of the Great Plains. From east to west, the High Plains rise in elevation from around 1,800 to 7,000 ft (550 to 2,130 m).[3]\\Thought 5: High Plains rise in elevation from around 1,800 to 7,000 ft, so the answer is 1,800 to 7,000 ft.\\Action 5: Finish[1,800 to 7,000 ft] \\ \\

Question: Musician and satirist Allie Goertz wrote a song about the "The Simpsons" character Milhouse, who Matt Groening named after who? \\ Thought 1: The question simplifies to \"The Simpsons\" character Milhouse is named after who. I only need to search Milhouse and find who it is named after.\\Action 1: Search[Milhouse]\\Observation 1: Milhouse Mussolini Van Houten is a recurring character in the Fox animated television series The Simpsons voiced by Pamela Hayden and created by Matt Groening.\\Thought 2: The paragraph does not tell who Milhouse is named after, maybe I can look up "named after".\\Action 2: Lookup[named after]\\Observation 2: (Result 1 / 1) Milhouse was named after U.S. president Richard Nixon, whose middle name was Milhous. \\Thought 3: Milhouse was named after U.S. president Richard Nixon, so the answer is Richard Nixon.\\Action 3: Finish[Richard Nixon] \\ \\

Question: Which documentary is about Finnish rock groups, Adam Clayton Powell or The Saimaa Gesture?\\ Thought 1: I need to search Adam Clayton Powell and The Saimaa Gesture, and find which documentary is about Finnish rock groups.\\Action 1: Search[Adam Clayton Powell]\\Observation 1: Could not find [Adam Clayton Powell]. Similar: ['Adam Clayton Powell III', 'Seventh Avenue (Manhattan)', 'Adam Clayton Powell Jr. State Office Building', 'Isabel Washington Powell', 'Adam Powell', 'Adam Clayton Powell (film)', 'Giancarlo Esposito'].\\Thought 2: To find the documentary, I can search Adam Clayton Powell (film).\\Action 2: Search[Adam Clayton Powell (film)]\\Observation 2: Adam Clayton Powell is a 1989 American documentary film directed by Richard Kilberg.\\The film is about the rise and fall of influential African-American politician Adam Clayton Powell Jr.[3][4] It was later aired as part of the PBS series The American Experience.\\Thought 3: Adam Clayton Powell (film) is a documentary about an African-American politician, not Finnish rock groups. So the documentary about Finnish rock groups must instead be The Saimaa Gesture.\\Action 3: Finish[The Saimaa Gesture] \\ \\

Question: What profession does Nicholas Ray and Elia Kazan have in common? \\ Thought 1: I need to search Nicholas Ray and Elia Kazan, find their professions, then find the profession they have in common.\\Action 1: Search[Nicholas Ray]\\Observation 1: Nicholas Ray (born Raymond Nicholas Kienzle Jr., August 7, 1911 - June 16, 1979) was an American film director, screenwriter, and actor best known for the 1955 film Rebel Without a Cause.\\Thought 2: Professions of Nicholas Ray are director, screenwriter, and actor. I need to search Elia Kazan next and find his professions.\\Action 2: Search[Elia Kazan]\\Observation 2: Elia Kazan was an American film and theatre director, producer, screenwriter and actor.\\Thought 3: Professions of Elia Kazan are director, producer, screenwriter, and actor. So profession Nicholas Ray and Elia Kazan have in common is director, screenwriter, and actor.\\Action 3: Finish[director, screenwriter, actor] \\ \\

Question: Which magazine was started first Arthur's Magazine or First for Women? \\ Thought 1: I need to search Arthur's Magazine and First for Women, and find which was started first.\\Action 1: Search[Arthur's Magazine]\\Observation 1: Arthur's Magazine (1844-1846) was an American literary periodical published in Philadelphia in the 19th century. \\Thought 2: Arthur's Magazine was started in 1844. I need to search First for Women next.\\Action 2: Search[First for Women]\\Observation 2: First for Women is a woman's magazine published by Bauer Media Group in the USA.[1] The magazine was started in 1989. \\Thought 3: First for Women was started in 1989. 1844 (Arthur's Magazine) $<$ 1989 (First for Women), so Arthur's Magazine was started first.\\Action 3: Finish[Arthur's Magazine] \\ \\

Question: Were Pavel Urysohn and Leonid Levin known for the same type of work? \\ Thought 1: I need to search Pavel Urysohn and Leonid Levin, find their types of work, then find if they are the same.\\Action 1: Search[Pavel Urysohn]\\Observation 1: Pavel Samuilovich Urysohn (February 3, 1898 - August 17, 1924) was a Soviet mathematician who is best known for his contributions in dimension theory.\\Thought 2: Pavel Urysohn is a mathematician. I need to search Leonid Levin next and find its type of work.\\Action 2: Search[Leonid Levin]\\Observation 2: Leonid Anatolievich Levin is a Soviet-American mathematician and computer scientist. \\Thought 3: Leonid Levin is a mathematician and computer scientist. So Pavel Urysohn and Leonid Levin have the same type of work. \\Action 3: Finish[yes] \\ \bottomrule
\multicolumn{1}{c}{\textbf{StrategyQA}}        \\ \toprule
\textbf{Standard} \\
Answer the question:\\
Question: Will the Albany in Georgia reach a hundred thousand occupants before the one in New York?\\Answer: No\\ \\
Question: Do the anchors on Rede Globo speak Chinese?\\Answer: No\\ \\ 
Question: Would a student of the class of 2017 have amnesia about 9/11?\\Answer: Yes\\ \\ 
Question: Is average number of peas in a pod enough commas for a billion?\\Answer: Yes \\ \hline
\textbf{Chain-of-Thought(CoT)} \\ 
Solve a question answering task. Your task is to generate Thought and Answer where a Thought can reason about the current situation by thinking step by step. \\
Here are some examples. \\
Question: Do the anchors on Rede Globo speak Chinese?\\Thought: The anchors on Rede Globo, a Brazilian television network, primarily speak Portuguese as that is the official language of Brazil. They may have proficiency in other languages, but Chinese is not likely to be one of the languages commonly spoken by the anchors on Rede Globo. So the answer is no.\\Answer: No\\ \\
Question: Will the Albany in Georgia reach a hundred thousand occupants before the one in New York?\\Thought: As of the most recent population estimates, Albany, New York, had a population of approximately 97,000 residents, while Albany, Georgia, had a population of around 73,000 residents. Albany, New York, is the capital of the state and is a major center for business, education, and government. It has a long history and economic significance, which attracts people to live and work in the area.On the other hand, Albany, Georgia, while an important regional center, is a smaller city in comparison. It does not have the same level of economic or cultural influence as Albany, New York. In conclusion, based on the current population figures and the different dynamics at play, it is unlikely that Albany, Georgia, will reach a population of one hundred thousand before Albany, New York. So the answer is no.\\Answer: No\\ \\
Question: Is average number of peas in a pod enough commas for a billion?\\Thought: Generally, a typical pea pod contains around 6 to 9 peas. A billion is a very large number 1,000,000,000 that requires 3 commas, which is less than the average number of peas in a pod. So the answer is yes.\\Answer: Yes\\ \\
Question: Is the language used in Saint Vincent and the Grenadines rooted in English?\\Thought: Saint Vincent and the Grenadines were once British colonies, and English became the dominant language during the colonial period. After gaining independence in 1979, English remained as the official language of the country, and it has continued to be used in education, government, media, and daily communication. English has permeated various aspects of society and is widely spoken by the population, though local dialects and accents may influence the spoken form of English in the region. So the answer is yes.\\Answer: Yes \\ \hline
\textbf{ReAct} \\
Solve a question answering task with interleaving Thought, Action, Observation steps. Thought can reason about the current situation, and Action can be three types: \\
(1) Search[entity], which searches the exact entity on Wikipedia and returns the first paragraph if it exists. If not, it will return some similar entities to search. \\ 
(2) Lookup[keyword], which returns the next sentence containing keyword in the current passage.
(3) Finish[answer], which returns the answer and finishes the task. \\
Here are some examples. \\
Question: Will the Albany in Georgia reach a hundred thousand occupants before the one in New York?\\Thought 1: I need to search Albany, Georgia and Albany, New York, find their current number of occupants, then compare to see which one will reach a hundred thousand first.\\Action 1: Search[Albany, Georgia]\\Observation 1: Albany is a city in the U.S. state of Georgia. Located on the Flint River, it is the seat of Dougherty County, and is the sole incorporated city in that county. The population was 77,434 at the 2010 U.S.\\Thought 2: The population of Albany, Georgia was 77,434 at the 2010 U.S.\\Action 2: Search[Albany, New York]\\Observation 2: Albany is the capital city of the U.S. state of New York and the seat of Albany County. It is located on the west bank of the Hudson River, about 10 miles (16 km) south of its confluence with the Mohawk River, and about 135 miles (220 km) north of New York City.\\Thought 3: It does not mention the population of New York City. So I need to look up population.\\Action 3: Lookup[population]\\Observation 3: (Result 1 / 1) As of 2020, Albany's population was 99,224.\\Thought 4: Albany, New York has 99,224 occupants in 2020, which is larger than Albany, Georgia. So Albany in Georgia will not reach a hundred thousand occupants before the one in New York, the answer is no.\\Action 4: Finish[No]\\ \\

Question: Do the anchors on Rede Globo speak Chinese?\\Thought 1: I need to search Rede Globo, and find if the anchors on Rede Globo speak Chinese.\\Action 1: Search[Rede Globo]\\Observation 1: TV Globo, or simply Globo and alternatively as Global), formerly known as Rede Globo, is a Brazilian free-to-air television network, launched by media proprietor Roberto Marinho on 26 April 1965.\\Thought 2: Rede Globo is a Brazilian television network, and Brazil is not a Chinese-speaking country, so anchors on Rede Globo do not speak Chinese.\\Action 2: Finish[No]\\ \\

Question: Would a student of the class of 2017 have amnesia about 9/11?\\Thought 1: I need to search the childhood amnesia and 9/11, and find if a student of the class of 2017 would have childhood amnesia about 9/11.\\Action 1: Search[childhood amnesia]\\Observation 1: Childhood amnesia, also called infantile amnesia, is the inability of adults to retrieve episodic memories (memories of situations or events) before the age of two to four years.\\Thought 2: Amnesia happens before the age of two to four years.\\Action 2: Search[9/11]\\Observation 2: The September 11 attacks, commonly known as 9/11, were four coordinated Islamist suicide terrorist attacks carried out by al-Qaeda against the United States in 2001.\\Thought 3: 9/11 attacks happened in 2001. A student of the class of 2017 is born around 1999, they would have been around 2 years old during the 9/11 attacks in 2001. Amnesia happens before the age of two to four years, so a student of the class of 2017 would have amnesia about 9/11.\\Action 3: Finish[Yes]\\ \\

Question: Is average number of peas in a pod enough commas for a billion?\\Thought 1: I need to search average number of peas in a pod and billion, and find if the average number of peas in a pod is enough commas for a billion.\\Action 1: Search[peas]\\Observation 1: The pea is most commonly the small spherical seed or the seed-pod of the flowering plant species Lathyrus oleraceus. Each pod contains several peas, which can be green or yellow.\\Thought 2: Each pod contains several peas. The number of several is usually three to five.\\Action 2: Search[billion]\\Observation 2: Billion is a word for a large number, and it has two distinct definitions: 1,000,000,000, i.e. one thousand million, or 109 (ten to the ninth power), as defined on the short scale.\\Thought 3: A billion is expressed as '1,000,000,000', which amounts to three commas. The average number of peas in a pod is three to five, which is indeed greater than the number of commas used in a billion (3 commas), so the answer is yes.\\Action 3: Finish[Yes] \\ \hline
\multicolumn{1}{c}{\textbf{MMLU}}                                                                  \\ \hline
\textbf{Standard} \\
Answer the question:\\
Question: A person takes buckets of water from the house and begins to add it to a pond in the yard. After a certain point, the pond\\A. breaks\\B. bloats\\C. sinks\\D. drowns\\Answer: B\\ \\

Question: Coal is solid rock that began as organic material that was deposited in a swamp. The formation of coal suggests that,\\A. coal is made mostly of skeletal remains of animals.\\B. coal is formed from magma that has solidified over time.\\C. it quickly becomes petrified when water is removed.\\D. geologic processes continue over millions of years.\\Answer: D\\ \\

Question: A student uses the following characteristics to describe a group of objects in space. * 200 billion stars * 30 million light years from Earth * 500 light years in diameter Which of the following is the student most likely describing?\\A. a galaxy\\B. the universe\\C. a constellation\\D. the solar system\\Answer: A\\ \\

Question: A microwave oven is connected to an outlet, 120 V, and draws a current of 2 amps. At what rate is energy being used by the microwave oven?\\A. 10 W\\B. 30 W\\C. 240 W\\D. 60 W\\Answer: C \\ \hline
\textbf{Chain-of-Thought(CoT)} \\ 
Solve a question answering task. Your task is to generate Thought and Answer where a Thought can reason about the current situation by thinking step by step. \\
Here are some examples. \\
Question: A person takes buckets of water from the house and begins to add it to a pond in the yard. After a certain point, the pond\\A. breaks\\B. bloats\\C. sinks\\D. drowns\\Thought: Each time the person adds a bucket of water, the level of water in the pond rises. Of all options, only B. bloats is consistent with the rise of water level. So the answer is B.\\Answer: B\\ \\

Question: Coal is solid rock that began as organic material that was deposited in a swamp. The formation of coal suggests that,\\A. coal is made mostly of skeletal remains of animals.\\B. coal is formed from magma that has solidified over time.\\C. it quickly becomes petrified when water is removed.\\D. geologic processes continue over millions of years.\\Thought: Let's evaluate each option. A. Coal is mostly composed of plant matter, not the skeletal remains of animals. Therefore, this option is incorrect. B. Coal is not formed from magma. Magma that solidifies over time creates igneous rocks, so this option is also incorrect. C. Petrification is a process by which organic material is turned into stone. It is not directly related to the process of coal formation, so this option is incorrect. D. The formation of coal takes incredibly long periods of time and consists of slow geologic processes such as sedimentation and metamorphism, which suggests that such processes continue over millions of years. Therefore, this option is correct. The answer is D.\\Answer: D\\ \\

Question: A student uses the following characteristics to describe a group of objects in space. * 200 billion stars * 30 million light years from Earth * 500 light years in diameter Which of the following is the student most likely describing?\\A. a galaxy\\B. the universe\\C. a constellation\\D. the solar system\\Thought: Let's evaluate each option. A. a galaxy: Possibly, as galaxies do contain billions of stars and can be millions of light years from Earth. B. the universe: Unlikely, as the universe is far larger than 30 million light years and contains more than just 200 billion stars. C. a constellation: Unlikely, as constellations are patterns of stars seen from Earth and don't have a physical size or distance associated with them. D. the solar system: Definitely not, as our solar system only contains one star, our sun. So, the answer is most likely A. a galaxy.\\Answer: A\\ \\

Question: A microwave oven is connected to an outlet, 120 V, and draws a current of 2 amps. At what rate is energy being used by the microwave oven?\\A. 10 W\\B. 30 W\\C. 240 W\\D. 60 W\\Thought: The rate energy is being used is equal to the power. For an electrical device P = IV, so P = (120)(2) = 240 W. So the answer is C.\\Answer: C \\ \hline
\textbf{ReAct} \\
Solve a question answering task with interleaving Thought, Action, Observation steps. Thought can reason about the current situation, and Action can be two types: \\
(1) search[question], which searches a question on Google and returns a short snippet containing the answer. Note that sometimes the snippet does not contain the answer, and some alternative search might be needed. \\
(2) finish[answer], which returns the answer and finishes the task. \\
Here are some examples. \\
Question: A person takes buckets of water from the house and begins to add it to a pond in the yard. After a certain point, the pond\\A. breaks\\B. bloats\\C. sinks\\D. drowns\\Thought: After continuously adding water to a pond, the pond will have more water than it could hold, thus bloats. So the answer is B.\\Action: finish[B]\\ \\

Question: Coal is solid rock that began as organic material that was deposited in a swamp. The formation of coal suggests that,\\A. coal is made mostly of skeletal remains of animals.\\B. coal is formed from magma that has solidified over time.\\C. it quickly becomes petrified when water is removed.\\D. geologic processes continue over millions of years.\\Thought: The question is about the formation of coal. I need to first learn how coal is formed.\\Action: search[How is coal formed?]\\Observation: Coal takes millions of years to form Coal contains the energy stored by plants that lived hundreds of millions of years ago in swampy forests. Layers of dirt and rock covered the plants over millions of years. The resulting pressure and heat turned the plants into the substance we call coal.\\Thought: Based on the information, I can check each option. A: coal is made by plants, not animals, so A is false. B: I have no information about if coal is formed from magma yet. I could search 'is coal formed from magma' to make sure. C: I have no information about if coal quickly becomes petrified when water is removed. I could search 'does coal quicklybecome petrified when water is removed' to make sure. D: Coal takes millions of years to form, so D is possibly true. I could search 'is the formulation of coal a geologic process' to make sure.\\Action: search[is the formulation of coal a geologic process]\\Observation: It is formed from plant remains that have been compacted, hardened, chemically altered, and metamorphosed by heat and pressure over geologic time.\\Thought: Seems the formulation of coal is over geologic time, so a geologic process. So the answer is D.\\Action: finish[D]\\ \\

Question: A student uses the following characteristics to describe a group of objects in space. * 200 billion stars * 30 million light years from Earth * 500 light years in diameter Which of the following is the student most likely describing?\\A. a galaxy\\B. the universe\\C. a constellation\\D. the solar system\\Thought: These options correspond to space systems of different sizes. I could search what is the diameter of each option to match.\\Action: search[what is the diameter of a galaxy]\\Observation: Most galaxies are 1,000 to 100,000 parsecs in diameter (approximately 3,000 to 300,000 light years) and are separated by distances on the order of millions of parsecs (or megaparsecs).\\Thought: A galaxy is usually 3,000 to 300,000 light years in diameter, which is slightly more than 500 light years. I should search the diameter of the universe next.\\Action: search[what is the diameter of the universe]\\Observation: 93 billion light-years\\Thought: The universe is 93 billion light years in diameter, which is much larger than 500 light years. I should search the diameter of a constellation next.\\Action: search[what is the diameter of a constellation]\\Observation: Its diameter, remarkably, is greater than 10 AU (1.5 billion kilometers!), large enough to fill the entire inner solar system almost as far out as Jupiter.\\Thought: A constellation is usually 10 AU in diameter. I need to convert it into light years.\\Action: search[10 AU to light years]\\Observation: 0.000158125\\Thought: A constellation is usually 0.000158125 light years in diameter, which is much smaller than 500 light years. I should search the diameter of the solar system next.\\Action: Sedna is three times farther away from Earth than Pluto, making it the most distant observable object known in the solar system. It is 143.73 billion km from the Sun, thus giving the Solar System a diameter of 287.46 billion km.\\Thought: The solar system is 287.46 billion km in diameter. I need to convert it into light years.\\Action: search[287.46 billion km to light years]\\Observation: 0.0303845459748716\\Thought: A constellation is usually 0.0303845459748716 light years in diameter, which is much smaller than 500 light years. Given all the information about diameters, the diameter of a galaxy is closest to 500 light years. So the answer is A.\\Action: finish[A]\\ \\

Question: A microwave oven is connected to an outlet, 120 V, and draws a current of 2 amps. At what rate is energy being used by the microwave oven?\\A. 10 W\\B. 30 W\\C. 240 W\\D. 60 W\\Thought: The question is about the rate of energy. I need to search what is the rate of energy.\\Action: search[what is the rate of energy]\\Observation: The rate at which energy is transferred is called power. When we say rate, this usually means a certain unit with respect to time or a unit over time. Power, as in the physics term, can be formally defined as the energy transferred per unit time.\\Thought: Given voltage and current, I need to search how to calculate power.\\Action: search[given voltage and current, how to calculate power]\\Observation: Power in Watts is calculated by multiplying voltage in Volts times current in Amps: 10 Amps of current at 240 Volts generates 2,400 Watts of power.\\Thought: In this case, the voltage is 120 volts and the current is 2 amps. Power=120V×2A=240W. So, the rate at which energy is being used by the microwave oven is 240 watts. So the correct answer is C. 240 W.\\Action: finish[C] \\ \bottomrule
\end{xltabular}

\end{document}